\documentclass[english]{article}

\usepackage{algorithmic}
\usepackage{algorithm}
\usepackage{amsfonts}
\usepackage{amsmath}
\usepackage{amssymb}
\usepackage{amsthm,wrapfig}
\usepackage{array}
\usepackage{arydshln}
\usepackage{bm}
\usepackage{cite}
\usepackage{color}
\usepackage{comment}
\usepackage{dsfont}
\usepackage{enumitem}
\usepackage{float}
\usepackage[T1]{fontenc}
\usepackage{geometry}
\usepackage{graphicx}
\usepackage{grffile}
\usepackage{hyperref}\hypersetup{colorlinks,linkcolor=blue,anchorcolor=blue,citecolor=blue}
\usepackage[latin9]{inputenc}
\usepackage{letltxmacro}
\usepackage{mathrsfs}
\usepackage{mathtools}
\usepackage{multirow}
\usepackage{nicefrac}
\usepackage{subcaption}
\usepackage{tablefootnote}
\usepackage{verbatim}

\usepackage{todonotes}

\usepackage{booktabs}

\allowdisplaybreaks

\newcommand{\ba}{\bm{a}}

\newcommand{\bk}{\bm{k}}

\newcommand{\bp}{\bm{p}}
\newcommand{\bq}{\bm{q}}

\newcommand{\bv}{\bm{v}}

\newcommand{\bz}{\bm{z}}

\newcommand{\bA}{\bm{A}}

\newcommand{\bH}{\bm{H}}

\newcommand{\bK}{\bm{K}}

\newcommand{\bV}{\bm{V}}
\newcommand{\bW}{\bm{W}}

\newcommand{\bDelta}{\bm{\Delta}}

\newcommand{\cD}{\mathcal{D}}

\newcommand{\cH}{\mathcal{H}}

\newcommand{\NN}{\mathbb{N}}

\newcommand{\RR}{\mathbb{R}}

\newcommand{\mfk}{\mathfrak} 
\newcommand{\one}{\bm{1}}
\newcommand{\zero}{\bm{0}}

\DeclareMathOperator{\bcdot}{\boldsymbol{\cdot}}

\allowdisplaybreaks
\makeatletter

\theoremstyle{plain} 

\theoremstyle{definition}
 
\theoremstyle{remark}

\definecolor{tian}{RGB}{0,150,0}
\definecolor{cm}{RGB}{250,0,200}
\definecolor{yc}{RGB}{255,0,0}
\definecolor{hd}{RGB}{0,180,200}

\definecolor{edits}{RGB}{250,100,100}

\DeclareMathOperator{\hho}{\text{H}_2 \text{O}}

\newcommand{\methodname}{\texttt{LESS}}

\begin{document}
\title{Get More with \texttt{LESS}: Synthesizing Recurrence with KV Cache Compression for Efficient LLM Inference}

 \author
 {
 	Harry Dong\thanks{Department of Electrical and Computer Engineering, Carnegie Mellon University, USA; Emails: \texttt{\{harryd,xinyuya2,yuejiec,beidic\}@andrew.cmu.edu}.} \\
	CMU 
	\and
 	Xinyu Yang\footnotemark[1]\\
    CMU 
    \and
 Zhenyu Zhang\thanks{Department of Electrical and Computer Engineering, University of Texas at Austin, USA; Emails:
 		\texttt{\{zhenyu.zhang,atlaswang\}@utexas.edu}.}  \\
 	 UT Austin   
    \and
    Zhangyang (Atlas) Wang\footnotemark[2]\\
    UT Austin 
    \and
 	Yuejie Chi\footnotemark[1] \\
 	CMU
    \and
    Beidi Chen\footnotemark[1] \thanks{Meta AI (FAIR), USA.} \\
 CMU \& Meta
 }

\date{\today}

\setcounter{tocdepth}{2}
\maketitle

\begin{abstract}

Many computational factors limit broader deployment of large language models. In this paper, we focus on a memory bottleneck imposed by the key-value (KV) cache, a computational shortcut that requires storing previous KV pairs during decoding. While existing KV cache methods approach this problem by pruning or evicting large swaths of relatively less important KV pairs to dramatically reduce the memory footprint of the cache, they can have limited success in tasks that require recollecting a majority of previous tokens. To alleviate this issue, we propose \methodname{}, a simple integration of a (nearly free) constant sized cache with eviction-based cache methods, such that all tokens can be queried at later decoding steps. Its ability to retain information throughout time shows merit on a variety of tasks where we demonstrate \methodname{} can help reduce the performance gap from caching everything, sometimes even matching it, all while being efficient. Relevant code can be found at \url{https://github.com/hdong920/LESS}.

\end{abstract}



\section{Introduction}
\label{sec:intro}

Throughout its lifetime, the transformer architecture \cite{vaswani2017attention} has made strides in natural language processing \cite{lin2022survey}, computer vision \cite{khan2022transformers}, healthcare \cite{nerella2023transformers}, and many other domains. Large language models (LLMs) \cite{zhang2022opt, scao2022bloom, fedus2022switch, anil2023palm, touvron2023llama2, team2023gemini, jiang2024mixtral} take transformers to the extreme by scaling the model, data, and context lengths to extraordinary levels. This has been remarkably useful for complex tasks such as chatbots, long document tasks, and biological sequences. However, during deployment, these tasks require generating long sequences or inputting large batch sizes, which places an immense computational burden on the key-value (KV) cache \cite{pope2023efficiently}, the storage of all previous keys and values at each layer to bypass recomputing them at future decoding steps. While this significantly saves computation, the tradeoff is an explosion of memory consumption. In such scenarios, the KV cache size often eclipses the model size. For instance, the Llama 2 7B model \cite{touvron2023llama2} occupies about 26 GB of memory, but the KV cache for an input of batch size 64 and sequence length 1024 occupies 64 GB of memory, nearly 2.5 times the model size. Hence, addressing this accessibility issue is imperative as LLMs continue to scale and break tight deployment constraints. 

\begin{figure}[t]
    \begin{center}
    \centerline{\includegraphics[width=0.55\columnwidth]{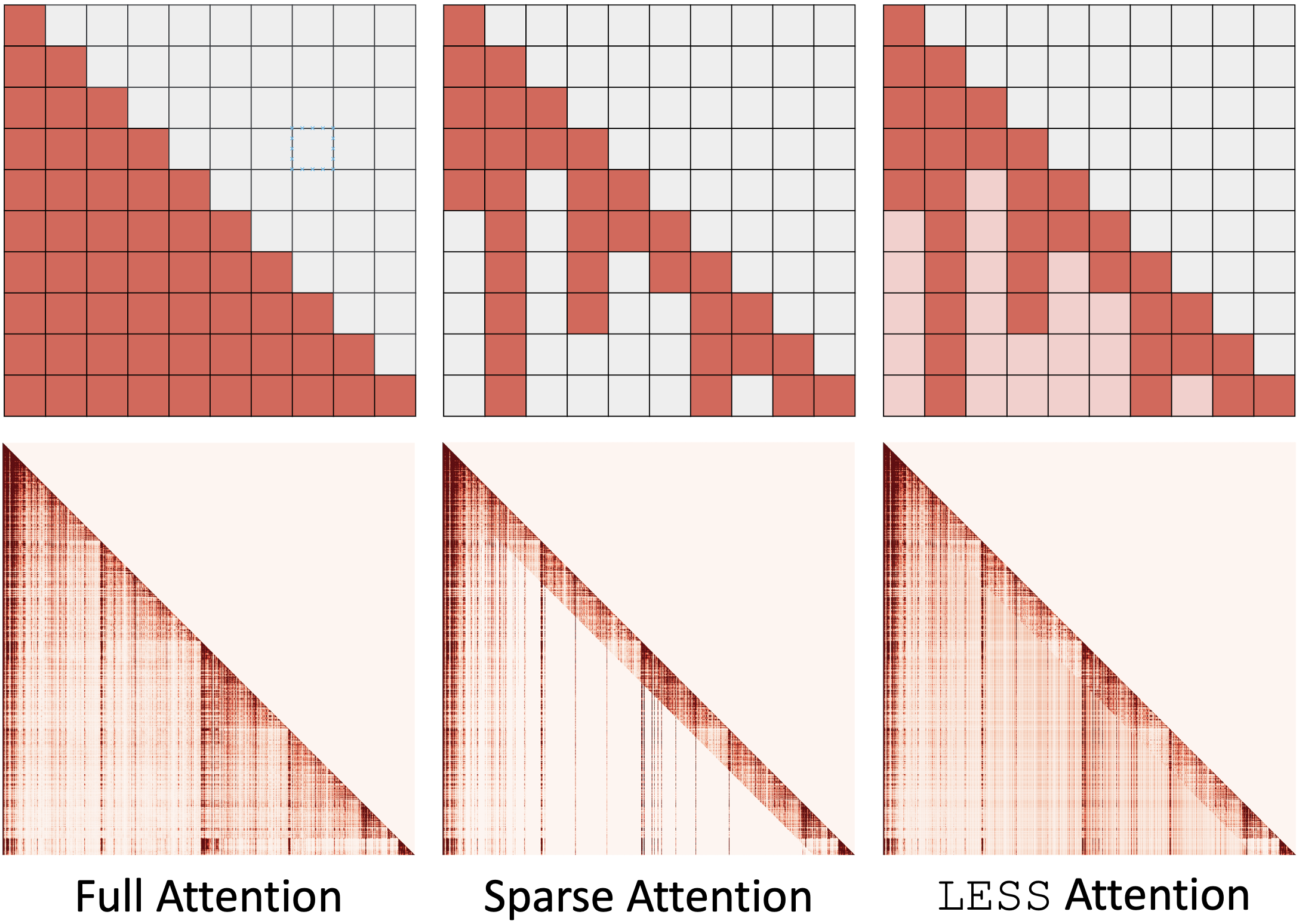}}
    \caption{Toy (top row) and Llama 2 7B (bottom row) example decoder attention maps with $\hho$ as the underlying sparse policy. In the top row, red/pink and grey squares are positive and zero attention probabilities, respectively. In the bottom row, darker colors indicate larger attention probabilities. Sparse attention policies zero out many positive attention probabilities. Our method, \methodname{}, ensures all previous tokens will have some contribution to the attention layer output to better retain information.}
    \label{fig:attn_maps}
    \end{center}
    \vskip -0.2in
\end{figure}

Thankfully, there have been initiatives to reduce the KV cache size. A line of work, in which we refer to as \textit{sparse policies or algorithms}, explores the selection of the best subset of KV pairs to cache \cite{zhang2023h,liu2023scissorhands,han2023lm, xiao2023efficient}. Although very promising, these methods are inevitably and irrecoverably discarding KV pairs deemed, in one way or another, less important than others, leading to gaps in attention maps 
\begin{wrapfigure}{r}{0.48\textwidth}
     \vskip -0.1in
    \begin{center}
\centerline{\includegraphics[width=0.45\columnwidth]{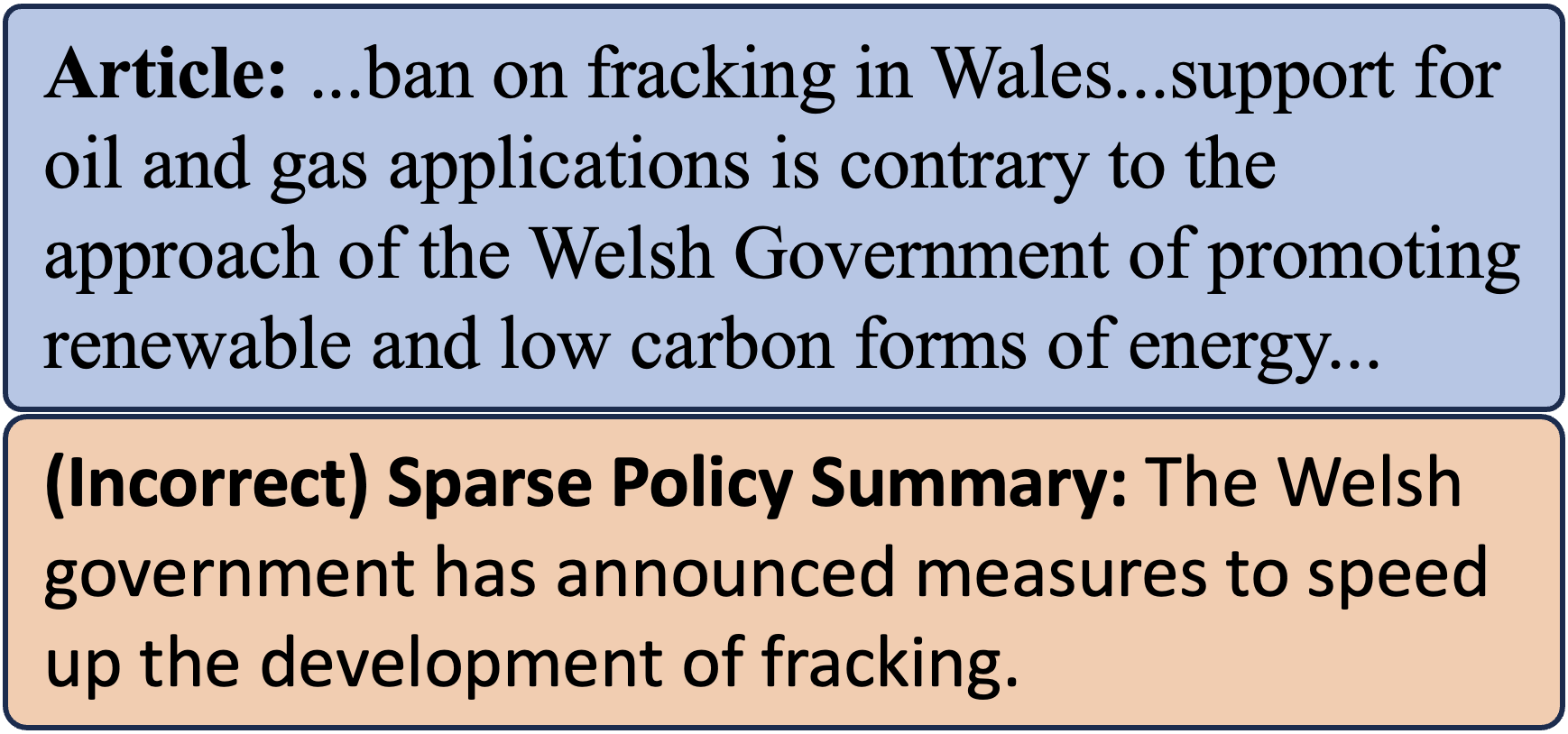}}
    \caption{Incorrect summary by Falcon 7B with sparse policy $\hho$.}
    \label{fig:wrong_summary}
    \end{center}
     \vskip -0.35in
\end{wrapfigure}
as shown in Figure \ref{fig:attn_maps}. 
Consequently, they are boldly assuming tokens that are unimportant now will not hold significance at future decoding steps, a faulty conjecture for tasks that deviate from this pattern. For instance, using  sparse policy $\hho$ \cite{zhang2023h} on Falcon 7B \cite{falcon40b} to summarize an article \cite{bbc2015fracking,narayan2018don} produces a \textit{factually incorrect} summary in Figure~\ref{fig:wrong_summary}. For the full article, see Figure~\ref{fig:example_falcon_out1} in Appendix~\ref{app:gen_outputs}.

One way to combat information loss is to cache more tokens, but this is far from memory efficient. An ideal KV cache policy should \textbf{1)} minimize performance degradation from a full cache, \textbf{2)} scale at a much slower rate than the full KV cache, and \textbf{3)} be cheap to integrate into existing pretrained LLMs. 

\begin{figure}[t]
    \centering
    \includegraphics[width=0.435\columnwidth]{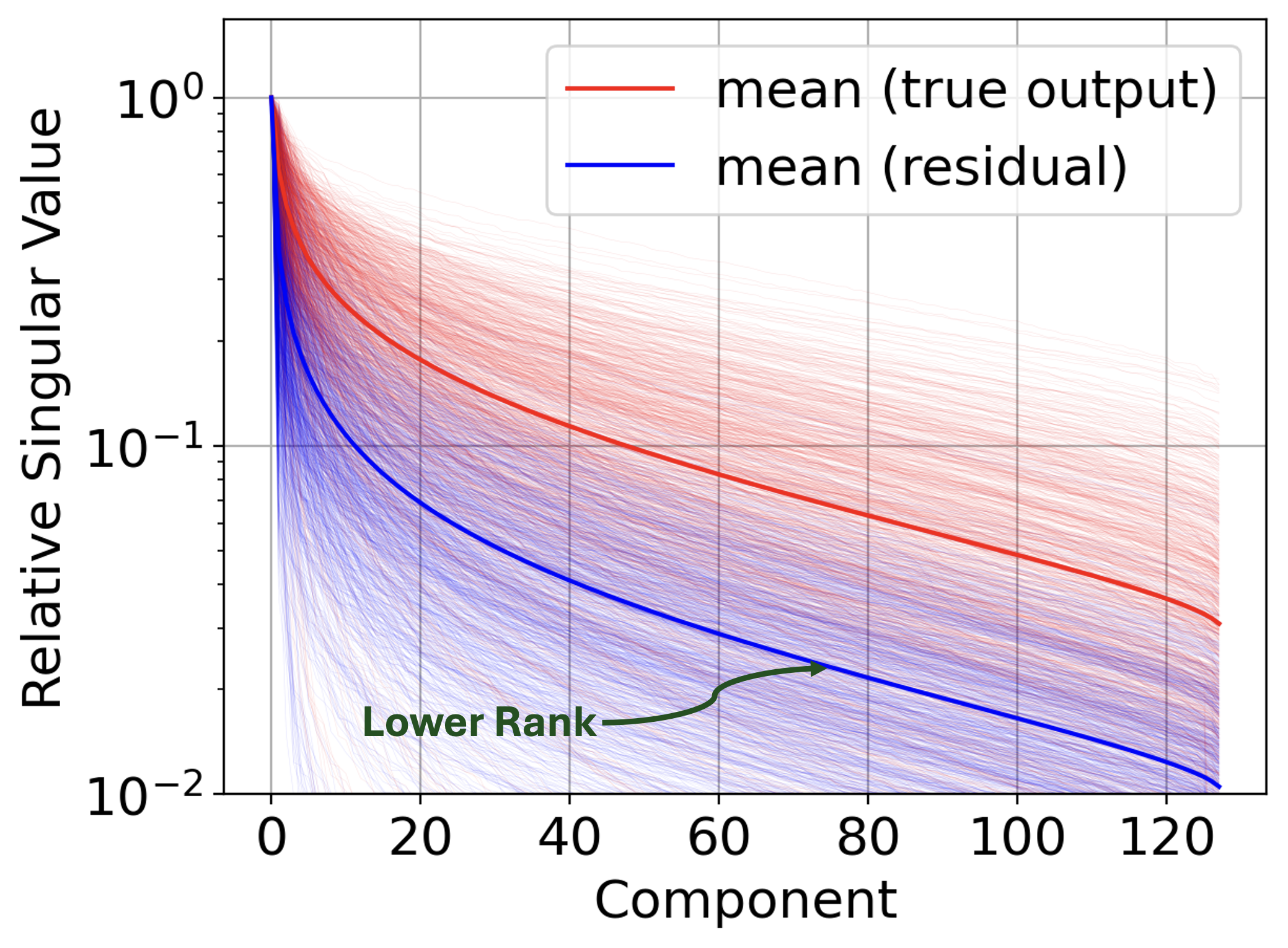}
    \includegraphics[width=0.42\columnwidth]{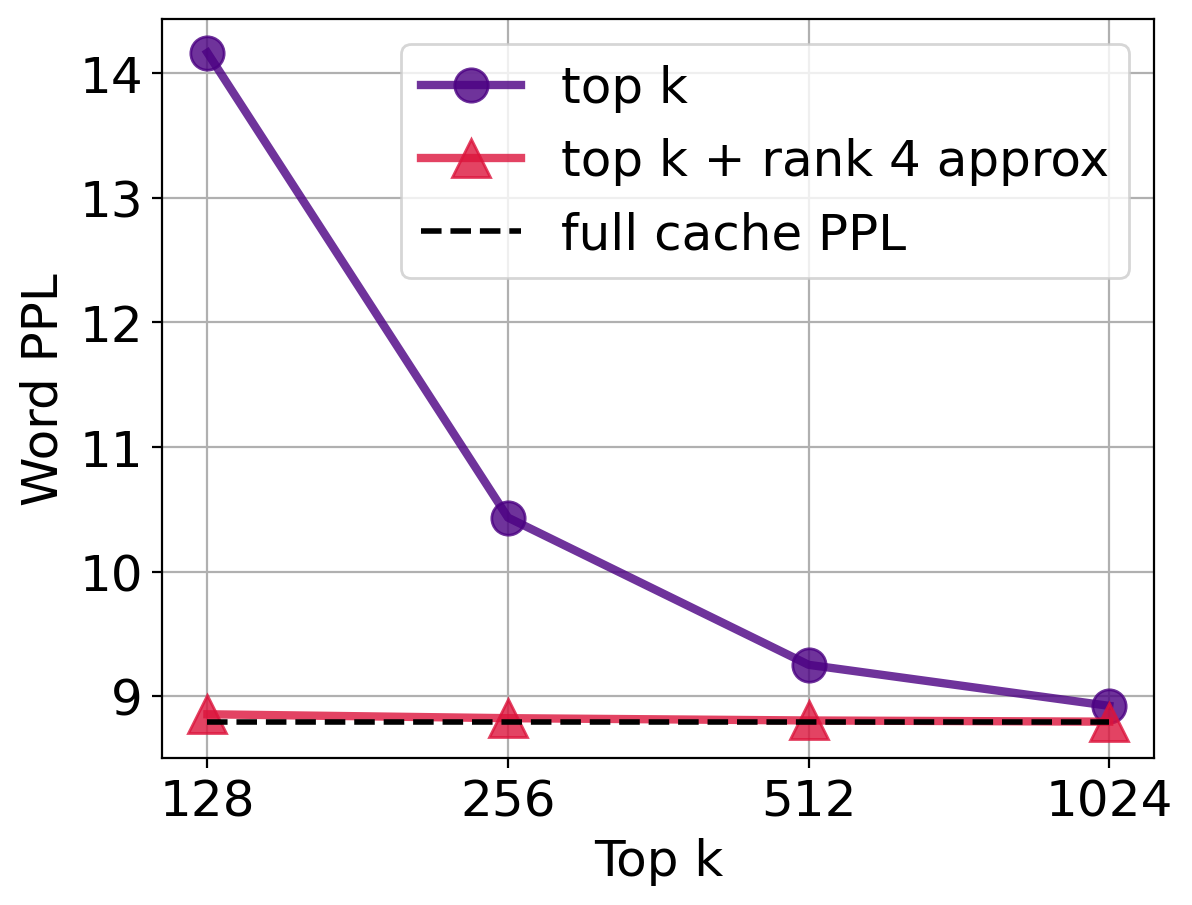}
    \caption{Attention residuals exploration in Llama 2 7B on WikiText \cite{merity2016pointer}. Mean and 1000 sample relative singular value plots of true attention outputs and residuals from top-$512$ sparse policy, showing the residual is much lower rank (left). End-to-end performance (lower is better) using top-$k$ caching with and without low-rank approximations (right). A rank-4 approximation virtually recovers the original performance.}
    \label{fig:residuals}
\end{figure}

Fortunately, with some investigation into the \textit{residual} between full and sparse attention outputs, a better strategy emerges. First, define the residual as $\bDelta_{\bA} = \bA - \bA_{\text{sparse}}$, where $\bA$ and $\bA_{\text{sparse}}$ are the full and sparse attention outputs, respectively. Using top-$k$ selection as our sparse policy, we observe the residuals $\bDelta_{\bA}$ are in fact \textit{low-rank} --- more so than $\bA$ --- based on Figure~\ref{fig:residuals}, a similar observation to Chen et al. \cite{chen2021scatterbrain}. Even a very low-rank approximation can nearly negate the performance degradation from sparse caching. \textit{In turn, this finding motivates the use of low-rank methods to approximate the residuals for efficient caches}.

We propose \methodname{} (\textbf{L}ow-rank \textbf{E}mbedding \textbf{S}idekick with \textbf{S}parse policy) to learn the \textit{residual} between the original attention output and the attention output approximated by a sparse policy. \methodname{} does this by accumulating information that would have been discarded by sparse policies into a constant-sized low-rank cache or state, allowing for queries to still access information to recover previously omitted regions in attention maps (see Figure \ref{fig:attn_maps}). 

We show that \methodname{} makes significant progress towards an ideal cache:
\begin{enumerate}
    \item \textbf{Performance Improvement:} \methodname{} synthesizes sparse KV policies with low-rank states to bridge the performance gap on a variety of tasks where these sparse algorithms show cracks of weakness. In fact, \methodname{} improves the performance much more than simply dedicating that memory to storing more KV pairs. 
    \item \textbf{Constant Low-rank Cache Size:} Low-rank caches in \methodname{} occupy constant memory with respect to the sequence length, and in our experiments, the extra storage to accommodate \methodname{} is \textit{nearly free}, taking up the equivalent space of only 4 extra KV pairs in our experiments. Inspired by recurrent networks, the low-rank state stores new information by recursive updates rather than concatenation. As each sample has its own cache, \methodname{} provides the same proportional cache reduction for small and large batch sizes.
    \item \textbf{Cheap Integration:} Changes to the LLMs' architectures are small and do not perturb the original weights. The only modifications to LLMs will be the addition of tiny multilayer perceptions (MLPs) at each attention layer. For example, using \methodname{} with Llama 2 13B adds fewer than 2\% of the total number of parameters. In addition, we can train \methodname{} at each attention layer independently from all others, bypassing expensive end-to-end training. Trained once, \methodname{} can transfer to more relaxed settings while maintaining comparable performance, further extending its applicability.
\end{enumerate}

Our comprehensive experiments on Llama 2 \cite{touvron2023llama2} and Falcon \cite{falcon40b} with different sparse policies \cite{zhang2023h, han2023lm, xiao2023efficient} on a variety of tasks demonstrates \methodname{} as a highly performative method that reduces KV cache memory. For instance, \methodname{} recovers more than 40\% of the Rouge-1 degradation caused by a sparse policy on the CNN/DailyMail dataset \cite{hermann2015teaching, see-etal-2017-get} with Falcon~7B. Finally, we provide an implementation of \methodname{} that
reduces the latency by up to $1.3\times$ and increases the throughput by $1.7\times$ from the full cache.

\paragraph{Notation.} We use unbolded letters (e.g. $a, A$), bold lowercase letters (e.g. $\ba$), bold uppercase letters (e.g. $\bA$) for scalars, row vectors, and matrices, respectively. Let $[\bA]_{i, \cdot}$ and $[\bA]_{\cdot, i}$ refer to the $i$-th row and column of $\bA$, respectively. Additionally, define $\zero_{n \times m}$ as a matrix of zeros and $\one_{n \times m}$ as a matrix of ones, both having shape $n \times m$. 

\section{Background \& Intuition}
\label{sec:background}

We start by building the intuition behind \methodname{}. Sparse and low-rank caches individually have noteworthy advantages but also severe drawbacks. Understanding the mechanisms of both (Section~\ref{sec:kv_policies} and \ref{sec:low_rank}) allows us to effectively synthesize sparse and low-rank structures to create \methodname{}. In Section~\ref{sec:sparse_low_rank}, we show that this type of synthesis is a principled approach which has also found success in other areas.

\subsection{KV Cache Policies}
\label{sec:kv_policies}

Many current methods to reduce the KV cache footprint involve keeping a tiny subset of the keys and values either with some pruning policy \cite{liu2023scissorhands, zhang2023h, han2023lm, xiao2023efficient, ge2023model, oren2024transformers} or a local attention mechanism \cite{child2019generating, parmar2018image}. The former method can be applied directly to trained models whereas the latter typically cannot, so with limited compute, deploying a KV cache pruning policy is more practical. Such methods take advantage of the observation that many tokens are irrelevant for attention in some tasks and thus omitting them leads to negligible performance loss. For instance, one of our baselines, $\hho$ \cite{zhang2023h}, continuously accumulates attention probabilities at each generation step to identify a set of heavy-hitting tokens to be cached together with the most recent tokens. Not explicitly designed for KV cache compression, algorithms for infinite inference \cite{han2023lm, xiao2023efficient} maintain a full cache, but as the input sequence exceeds the maximum context length of a model, KV pairs in the middle of the sequence are dropped. Staying within the maximum context length, this results in a cache that maintains the most recent and first few tokens. Regardless of the sparse method, maintaining a tight KV cache budget can seriously impair model performance, as we will see in Section~\ref{sec:experiments}.

There also exist promising non-eviction based methods. DMC involves using a large amount of data to fine tune models to choose between accumulating or appending each KV pair \cite{nawrot2024dynamic}.
CacheGen's KV cache compression at the bit-level takes a query-agnostic approach \cite{liu2023cachegen}. 
In vision tasks, token merging is an effective way to cut down the number of tokens to process \cite{bolya2022token, renggli2022learning}.

\subsection{Low-rank Attention}
\label{sec:low_rank}

Low-rank structures in attention have been explored extensively \cite{tay2022efficient}, namely from the lens of recurrent neural networks (RNNs). Unlike transformers, RNNs integrate information from all previous tokens into \textit{hidden states}, analogous low-rank structures to KV caches that organically occupy constant memory. In fact, this feature of RNNs over transformers has motivated research in alternative architectures \cite{dao2022hungry,poli2023hyena,peng2023rwkv,sun2023retentive, gu2023mamba}, but for now, their adoption in LLMs is very limited compared to transformers. Though not as performative as these alternative architectures, linear transformers that break apart the attention operation into kernels also maintain a constant sized KV cache \cite{tsai2019transformer, katharopoulos2020transformers, choromanski2020rethinking, peng2021random} by reformulating the cache into an RNN hidden state. These types of caching mechanisms are \textit{low-rank} since information is condensed along the sequence axis, rather than explicitly maintaining individual tokens. 
This is possible when we replace the softmax with a separable similarity metric $\phi (\bq_t) \psi (\bK_t)^\top$ for some row-wise functions $\phi$ and $\psi$, letting $\bq_t \in \RR^{1 \times D}$ and $\bK_t \in \RR^{t \times D}$ be the query and keys at step $t$, respectively. To elaborate, if $\phi$ and $\psi$ are such that
\begin{align*}
    \ba_t = \text{softmax}\left(\frac{\bq_t \bK_t^\top}{\sqrt{D}}\right) \bV_t \approx \frac{\phi (\bq_t) \psi (\bK_t)^\top \bV_t}{\phi (\bq_t) \psi (\bK_t)^\top \one_{S \times 1}}, 
\end{align*}
we just need to cache hidden states $\bH_t = \psi (\bK_t)^\top \bV_t \in \RR^{R \times D}$ and $\bz_t = \sum_{s= 1}^t \psi ([\bK_t]_s) \in \RR^{1 \times R}$ for inference which can be expressed recursively as 
\begin{align*}
    \bH_{t+1} &= \bH_{t} + \psi (\bk_t)^\top \bv_t, \\
    \bz_{t+1} &= \bz_{t} + \psi (\bk_{t})
\end{align*}
for each new KV pair $(\bk_t, \bv_t)$. At initialization, $\bH_0 = \zero_{R \times D}$ and $\bz_0 = \zero_{1 \times R}$. This is a clear improvement from having to store ever increasing sizes of $\bK_t$ and $\bV_t$, as the memory consumption is independent from $t$. Note that our presentation differs slightly  since we do not constrain $\phi = \psi$ \cite{chen2023primal}. With this formulation, transformers act like RNNs which occupy constant memory during generation by not appending but updating hidden states during each generation step. Since LLMs are not typically trained in this fashion, a major challenge is to induce this property without significant computation or adjustment to the original weights \cite{kasai2021finetuning}. While its dilution of information restricts its performance when specific tokens need to be recalled with strong signals \cite{khandelwal2018sharp}, this is exactly what a sparse KV cache algorithm can do, so we can fully take advantage of its infinite compression capability to obtain some high level representation of the less important tokens, meaning kernelized attention is a good candidate method for \methodname{} to learn the residual.

\subsection{Sparse and Low-rank Decomposition}
\label{sec:sparse_low_rank}

\methodname{} follows a rich history of decomposing structures into sparse and low-rank components. 
Particularly, the study of robust principal component analysis \cite{candes2011robust, chandrasekaran2011rank} has shown this type of decomposition greatly enhances approximation accuracy and expressibility beyond just sparse or low-rank matrices alone. Its success has spread to deep learning areas such as efficient attention \cite{chen2021scatterbrain}, model compression \cite{li2023losparse}, and fine-tuning \cite{nikdan2024rosa}. Likewise, we take inspiration from these works in our design.

\section{Method}
\label{sec:method}

\begin{figure*}[t]
\begin{center}
\centering
\includegraphics[width=\linewidth]{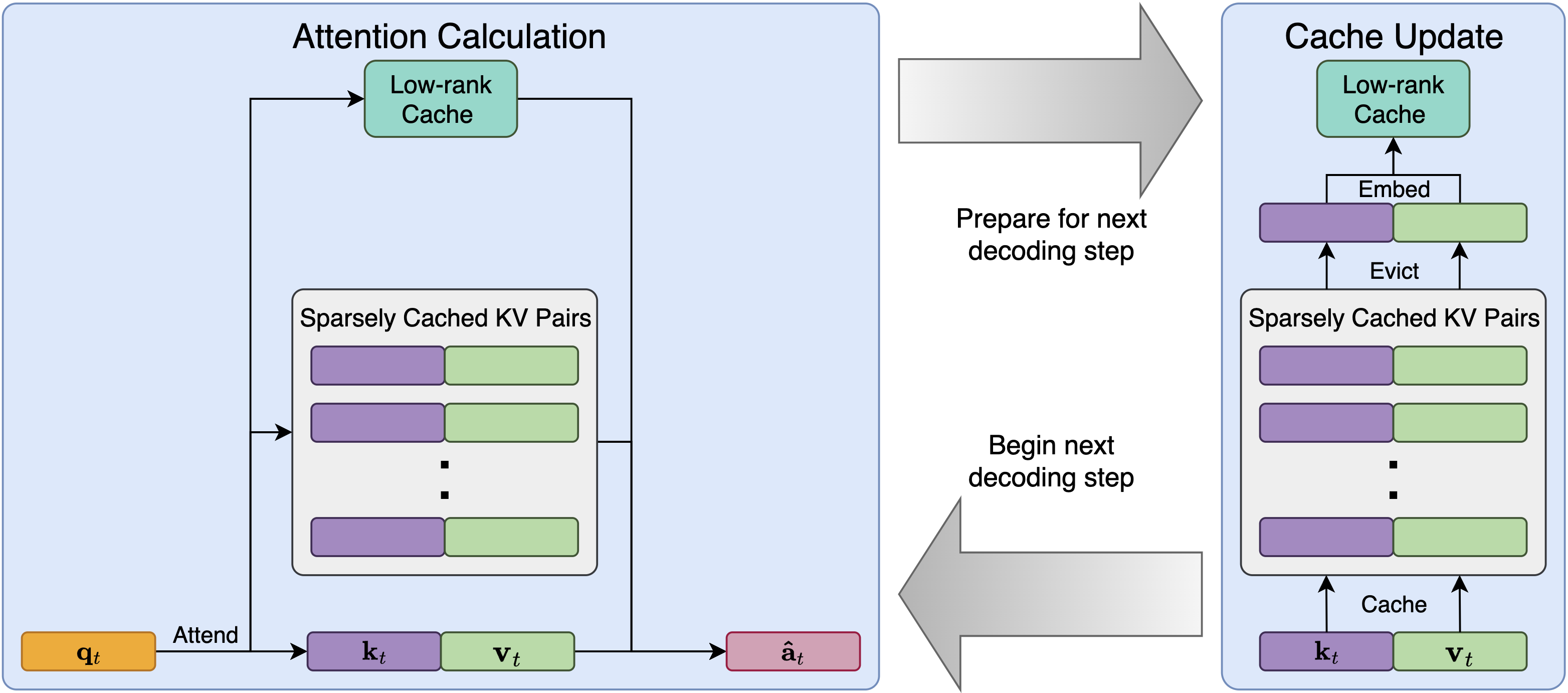}\par
\caption{\methodname{} algorithm during inference. At each decoding step, attention is calculated as in \eqref{eq:attn_approx}. To prepare for the next decoding step, the cache is updated by placing the most recent KV pair into the sparse policy cache, and if it has exceeded capacity, a KV pair will be evicted and integrated into the low-rank cache $\bH_t$ before being deleted.}
\label{fig:algorithm}
\end{center}
\end{figure*}

When we convert the intuition in Section~\ref{sec:background} into an algorithm, a couple technical challenges arise. One challenge is finding an effective way to mix attention probabilities produced by sparse policies and low-rank kernels. Second, we need to design a framework general enough to work with a broad class of sparse policies. In some cases, different sparse policies may be preferable, so our method should be compatible with many sparse policies. Third, our method should be cheap compute to develop. We show that \methodname{} overcomes all these challenges in a two step process: attention computation followed by cache updates.

\subsection{KV Caching with \methodname{}}

We propose \methodname{}, a general method to synthesize low-rank caches with \textit{any} eviction-based sparse KV cache policy, $\mfk{C}$, to close the performance gap from full KV caching while being efficient. Notably, our method only adds a constant sized cache which does not scale with the sequence length. For the sparse policy, $\mfk{C}$, we require that it can output the cached keys $\bK_{\mfk{C},t} \in \RR^{B_t \times D}$, the cached values $\bV_{\mfk{C},t} \in \RR^{B_t \times D}$, and the set of discarded KV pairs $\cD_t$ at iteration $t$ where $B_t \in \NN$ is the number of cached pairs. 

Letting $\bcdot$ denote both $\phi$ and $\psi$, we define our kernels as 
\begin{align}
    \phi(\bq) &= \left| \sigma_\phi(\sigma_\phi(\bq \bW_{\phi, 1}) \bW_{\phi, 2}) \right| \label{eq:kernel_q}\\
    \psi(\bk) &= \left| \sigma_\psi(\sigma_\psi(\bk \bW_{\psi, 1}) \bW_{\psi, 2}) \bW_{\psi, 3} \right| \label{eq:kernel_k}
\end{align}
for activation functions $\sigma_{\bcdot}$, $\bW_{\bcdot, 1} \in \RR^{D \times R'}$, $\bW_{\bcdot, 2} \in \RR^{R' \times R}$, and $\bW_{\psi, 3} \in \RR^{R \times R}$. The element-wise absolute values ensure the inner product $\phi(\bq) \psi(\bk)^\top > 0$ to preserve the nonnegativity of attention probabilities. In the ideal case, if $\phi(\bq) \psi(\bk)^\top = e^{\bq \bk^\top / \sqrt{D}}$ for all $\bq, \bk$, then the result would be the original attention probabilities. 

\paragraph{Attention Calculation.} Now, we describe the attention calculation procedure summarized in Algorithm~\ref{alg:main}. At step $t$, we find the query-key-value triplet $(\bq_{t}, \bk_{t}, \bv_{t})$ from the input token as usual. Recalling that we have cached $\bK_{\mfk{C},t}$, $\bV_{\mfk{C},t}$, $\bH_t$, and $\bz_t$ from the previous generation step, append $\bk_{t}$ to $\bK_{\mfk{C},t}$ and $\bv_{t}$ to $\bV_{\mfk{C},t}$ to obtain $\bK_{\mfk{C},t}^\prime \in \RR^{(B_t+1) \times D}$ and $\bV_{\mfk{C},t}^\prime \in \RR^{(B_t+1) \times D}$, respectively. Then, we can find $\hat{\ba}_{t}$, our approximation of the original attention $\ba_{t}$, by computing
\begin{align}\label{eq:attn_approx}
    \hat{\ba}_{t} = \frac{\phi (\bq_{t}) \bH_t + \exp(\bq_{t} (\bK_{\mfk{C},t}^\prime)^\top / \sqrt{D}) \bV_{\mfk{C},t}^\prime}{\phi (\bq_{t}) \bz_t^\top + \exp(\bq_{t} (\bK_{\mfk{C},t}^\prime)^\top / \sqrt{D}) \one_{B \times 1}} .
\end{align}
During the prompting phase (i.e. $t=0$), it is just regular attention since $\bH_0 = \zero_{R \times D}$ and $\bz_0 = \zero_{1 \times R}$.

\paragraph{Cache Updates.} With the attention computed, we need to prepare the necessary ingredients for iteration $t+1$ by finding $\bK_{\mfk{C},t+1}$, $\bV_{\mfk{C},t+1}$, $\bH_{t+1}$, and $\bz_{t+1}$. The first two are simple since the sparse policy will return $\bK_{\mfk{C},t+1}$, $ \bV_{\mfk{C},t+1}$, and $\cD_{t+1}$. Before freeing $\cD_{t+1}$ from memory, we embed its information into $\bH_{t+1}$ and $\bz_{t+1}$:
\begin{align}
    \bH_{t+1} &= \bH_t + \sum_{(\bk, \bv) \in \cD_{t+1}} \psi (\bk)^\top \bv, \label{eq:h_update}\\
    \bz_{t+1} &= \bz_t + \sum_{(\bk, \bv) \in \cD_{t+1}} \psi (\bk) . \label{eq:z_update}
\end{align}
After this, $\cD_{t+1}$ can be deleted, and we are prepared for the following generation step. Intuitively, $\bH_{t+1}$ and $\bz_{t+1}$ are updated recursively by keys and values that have been newly pruned at each decoding step. As such, they are constant size repositories of information from all deleted KV pairs which becomes clear when we expand the recursion:
\begin{align}
    \bH_{t+1} &= \sum_{(\bk, \bv) \in \bigcup_{i=1}^{t+1} \cD_{i}} \psi (\bk)^\top \bv, \label{eq:h_update_it}
\end{align}
and similarly for $\bz_{t+1}$.

\begin{algorithm}[tb]
\caption{Generation Step with \methodname{}}\label{alg:main}
\begin{algorithmic}

\STATE {\bfseries Input:} $\mfk{C}, \bq_{t}, \bk_{t}, \bv_{t}$

\STATE Load $\bK_{\mfk{C},t}, \bV_{\mfk{C},t}, \bH_t, \bz_t$ from memory.
\STATE $\bK_{\mfk{C},t}^\prime \gets \text{concatenate}(\bK_{\mfk{C},t}, \bk_{t})$
\STATE $\bV_{\mfk{C},t}^\prime \gets \text{concatenate}(\bV_{\mfk{C},t}, \bv_{t})$
\STATE Obtain $\hat{\ba}_{t}$ via \eqref{eq:attn_approx}.
\STATE Obtain $\bK_{\mfk{C},t+1}, \bV_{\mfk{C},t+1}, \cD_{t+1}$ from sparse KV cache algorithm $\mfk{C}$.
\STATE Update $\bH_{t+1}$ via \eqref{eq:h_update}.
\STATE Update $\bz_{t+1}$ via \eqref{eq:z_update}.
\STATE Save $\bK_{\mfk{C},t+1}, \bV_{\mfk{C},t+1}, \bH_{t+1}, \bz_{t+1}$.
\STATE Delete $\cD_{t+1}$.
\STATE {\bfseries Return:} $\hat{\ba}_{t}$
\end{algorithmic}
\end{algorithm}

\subsection{Implementation Details}

\paragraph{Inexpensive Training.} With our inference-time protocol outlined, we now describe how we can cheaply train our kernel functions $\phi$ and $\psi$. Because training end-to-end is time consuming and resource intensive, we choose to  train $\phi$ and $\psi$ at each layer independent of all other layers which already surprisingly gives strong results. The training objective is to minimize the $\ell_2$ distance to the output projection of the original attention layer using that layer's inputs. All weights except for those in $\phi$ and $\psi$ are frozen. As a result, the only computational requirements are the abilities to backpropagate through a single attention layer and run inference on the full model to collect a dataset of attention layer inputs and outputs, which for all models we experiment with, \textit{can be done on a single GPU}. With more devices, training each layer can be parallelized. While inference follows recursive updates of $\bH_t$ and $\bz_t$, this does not impede parallelism along the sequence axis because we can just construct the full attention matrix where entries not computed by sparsely cached KV pairs, as determined by whichever sparse policy we train on, will be found by the kernel functions. 

All training runs used identical hyperparameters for simplicity. \methodname{} was trained using Adam \cite{kingma2014adam} for 40 epochs with an initial learning rate of 0.001 which halved every 10 epochs. 
We fixed the hidden layer dimension $R' = 512$, used a dropout rate of 0.3 within the kernels, and let all nonlinear functions $\sigma_\phi$ and $\sigma_\psi$ to be GELUs. None of the original model's weights are updated. First, we sampled 512 sequences for Llama 2 models \cite{touvron2023llama2} and 1024 sequences for Falcon \cite{falcon40b} from the C4 training set \cite{2019t5}. Since Falcon's context length is half of Llama 2's, the training sets have the same number of tokens. Next, queries, keys, and values at each layer would be collected as each sample propagated through the models. These collected features (fed in batches of 2) would be used to train the kernels at each layer independently using some sparse policy at some sparsity level. For multi-query attention \cite{shazeer2019fast}, we extend $\hho$ to aggregate attention scores across all query attention heads to determine KV pairs to evict.

We find that the kernel initialization is critical. As we will show in our experiments (Section \ref{sec:experiments}), the sparse policies already have decent performance which we want to use as a starting point. As such, we add learnable scalars between layers in $\psi$ which are initially set to $10^{-4}$, so the influence of \methodname{} during the first few gradient steps is small. In this way, the sparse policy acts as a warm start, and we can immediately reduce the sparse policy's residual.

\paragraph{Efficient Generation.} We also develop an implementation that enhances throughput and reduces the latency of LLM generation of \methodname{}. For the sparse cache, we adapt the implementation from Zhang et al. \cite{zhang2023h} to support any KV cache eviction algorithm efficiently. To avoid data movement in memory, we directly replace the evicted KV pair with the newly-added one. As our kernels are small MLPs with GELUs, we implement a fused linear kernel that absorbs the activation into the layer before to avoid writing the intermediate results to DRAM for the low-rank cache. 
\vspace{-0.1in}

\section{Experiments}
\label{sec:experiments}


Here, we demonstrate the impressive performance of \methodname{} across multiple datasets, models (Llama 2 and Falcon), sparse policies \cite{zhang2023h, han2023lm, xiao2023efficient, oren2024transformers}, and sparsity levels, despite allocating only approximately 4 tokens of storage to the low-rank state. In Section~\ref{sec:lm}, \methodname{} achieves the closest performance to the full cache in language modeling and classification tasks. For example, evaluated with $2\% \hho$ in Llama 2 7B, \methodname{} reduces the word perplexities on WikiText and PG-19 by over 20\% from $\hho$ alone, relative to the full cache performance. Section~\ref{sec:summarization} shows similar gains in summarization. For example, \methodname{} reduces Rouge-1 degredation by $10\% \hho$ in Falcon 7B on CNN/DailyMail by 41.4\%. In Section~\ref{sec:efficiency}, we note the lower latency ($1.1 - 1.3\times$ reduction) and higher throughput of \methodname{} ($1.7\times$ higher) compared to full caching. Finally, in Section~\ref{sec:ablation}, we discuss different characteristics of \methodname{}, namely the recovery of true attention probabilities, kernel size scaling, and capabilities for long sequences. 

We explore three sparse policies: $\hho$ \cite{zhang2023h}, $\Lambda$-masking from the infinite generation literature \cite{han2023lm, xiao2023efficient}, and TOVA \cite{oren2024transformers}. When using $\hho$, the sparse KV cache is equally split between the heavy hitter tokens and the recent tokens (e.g. 5\% $\hho$ cache consists of 2.5\% heavy hitters and 2.5\% recent tokens). For $\Lambda$-masking, the cache consists of the first 4 and most recent tokens. The percentages represent how much of the model's max context length is cached, so regardless of input length, the cache size remains the same for fairness. Since the sparsity level can translate to different numbers of tokens among models based on the max input lengths, we include Table \ref{tab:model_details} as a quick reference for the models we evaluate on, Llama 2 and Falcon. The token count is rounded down to the nearest even number to make sure $\hho$ can have an even split.  

\begin{table}[t]
\caption{Token counts at different sparsity levels.}
\begin{center}
\begin{small}
\begin{sc}
\begin{tabular}{lcc}
\toprule
Model & Max Length & \# Tokens at 2\%/5\%/10\% \\
\midrule
Llama 2 & 4096 & 80 / 204 / 408  \\
Falcon & 2048 & 40 / 102 / 204 \\
\bottomrule
\end{tabular}
\end{sc}
\end{small}
\end{center}
\label{tab:model_details}
\end{table}

\begin{figure}[t]
\begin{center}
\centerline{\includegraphics[width=0.6\columnwidth]{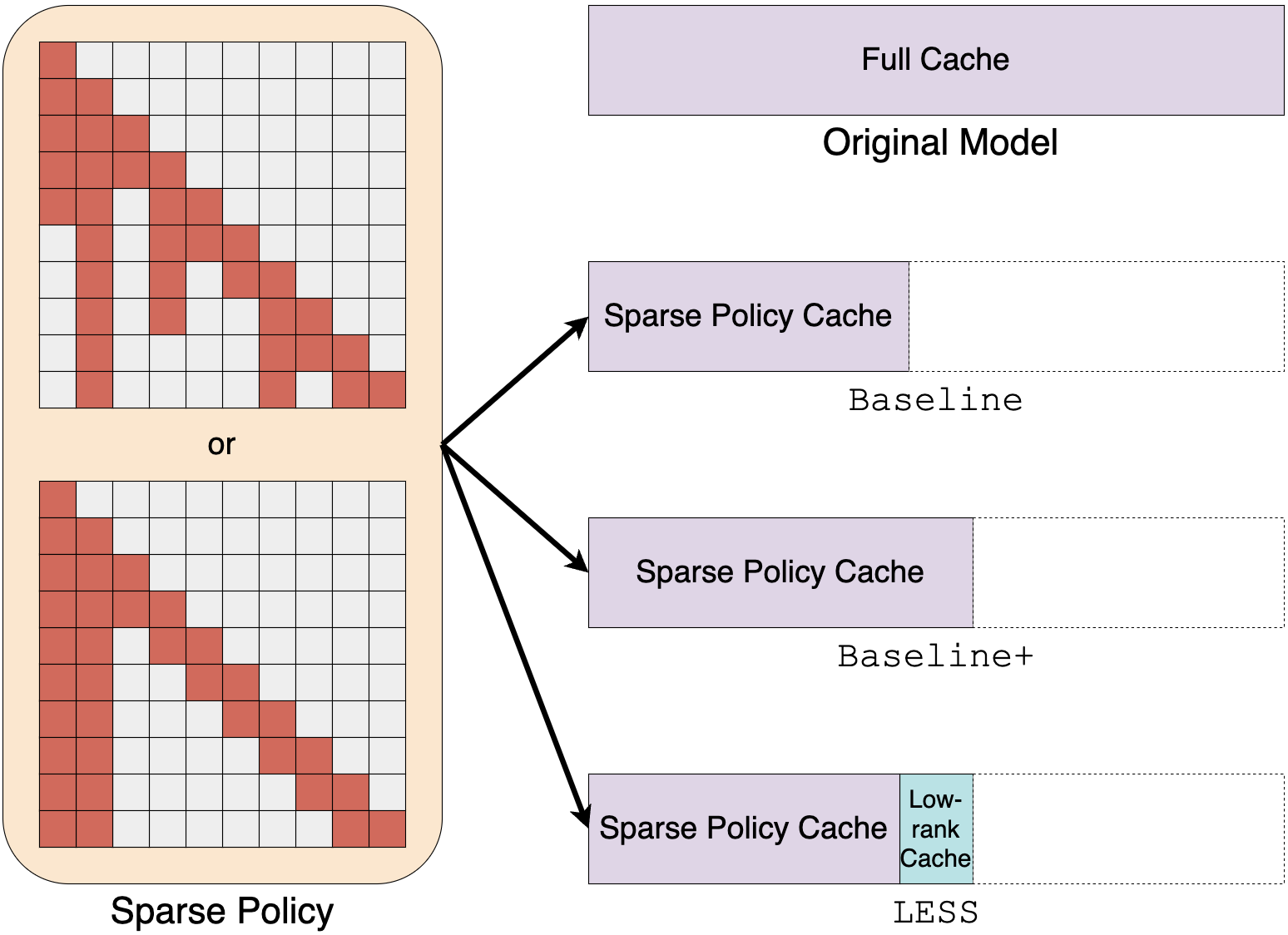}}
\caption{Experimental setup. First, a sparse policy is chosen as the underlying policy behind all methods. Then, we compare performance among the full cache model, \texttt{Baseline}, \texttt{Baseline+}, and  \methodname{}. \texttt{Baseline+} and \methodname{} use the same amount of storage which is slightly larger than the requirements of \texttt{Baseline}.}
\label{fig:baseline_setup}
\end{center}
 \vskip -0.2in
\end{figure}

For our experiments, we set the kernel size $R=8$, unless otherwise stated. Thus, while minuscule, the size of $\bH$ is nonzero, equivalent to caching 4 extra tokens. We ignore the influence of $\bz$ since it only has $R$ entries. As such, when evaluating on a task at $\alpha$\% sparsity, we compare \methodname{} with the sparse policy $\mfk{C}$ at $\alpha$\% sparsity and at $\alpha$\% sparsity plus additional tokens to match the extra space taken by $\bH$ (e.g. 4 tokens in experiments where $R=8$), which we denote as \texttt{Baseline} and \texttt{Baseline+}, respectively. Both are inherently sparse-only policies. A visual representation of the different baselines can be found in Figure \ref{fig:baseline_setup}. Note that the sparsity level and policy $\mfk{C}$ will vary throughout our experiments depending on the context. The purpose of evaluating \texttt{Baseline} is to compare the performance gain from extra tokens and the low-rank state $\bH$. Additionally, we evaluate the full KV cache to observe how far we are from the unconstrained potential of the original model. For our method, we denote it as \methodname{} ($\beta$\%) where $\beta$ is the percent cache sparsity \methodname{} was trained on with some sparse policy depending on the context.

\subsection{Language Modeling \& Classification}
\label{sec:lm}

We start with validating our method trained at different sparsity levels on some language modeling and classification tasks at different sparsity levels using Language Modeling Evaluation Harness \cite{eval-harness}. For these tasks, we use the same setup as in training by masking out query-key interactions depending on the sparse policy and having \methodname{} capture the masked probabilities. In addition, we purposefully mismatch training and testing sparsity levels to uncover insight on the transferability between test sparsity levels. To illustrate why a learned kernel is necessary, we also evaluate $\hho$ with Performer kernels \cite{choromanski2020rethinking} based on random Fourier features \cite{rahimi2007random}, which we denote as $\hho$+Performer.

Table \ref{tab:llama_lm} shows Llama 2 7B performance on WikiText \cite{merity2016pointer} and PG-19 \cite{raecompressive2019, gao2020pile} using $\hho$. Looking at the scenarios where training sparsity is equal to the test sparsity, our method is able to achieve much lower word perplexities than the baselines. Notably, \methodname{} beats \texttt{Baseline} by a wider margin than \texttt{Baseline+} and $\hho$+Performer, indicating that \methodname{} uses the space of 4 extra tokens most effectively. The lackluster performance of $\hho$+Performer suggests that learned kernels are needed to make a noticeable improvement. Moreover, \methodname{} trained at one sparsity level can often generalize reasonably to higher sparsity levels especially on WikiText, even sometimes matching the performance of ones trained at the test sparsity level. The reverse is less effective but can still be better than the baselines. However, all methods are still quite far from the full cache performance. 





\begin{table}[t]
\caption{Llama 2 7B WikiText and PG-19 word perplexities with $\hho$ as the primary underlying sparse policy. Numeric column names indicate the sparsity levels during test time. Lower is better.}
\begin{center}
\begin{small}
\begin{sc}
\begin{tabular}{lccc}
\toprule
$\hho$ Method & 2\% $\hho$ & 5\% $\hho$ & 10\% $\hho$ \\
\midrule

\textit{WikiText} \\ 
Full Cache & 8.791 & 8.791 & 8.791 \\
\texttt{Baseline} & 13.333 & 9.863 & 9.295 \\
\texttt{Baseline+} & 12.718 & 9.842 & 9.288 \\
$\hho$+Performer & 13.332 & 9.863 & 9.296 \\
\methodname{} (2\%) & \textbf{10.745} & 9.658 & 9.261 \\
\methodname{} (5\%) & 11.321 & \textbf{9.657} & 9.239 \\
\methodname{} (10\%)  & 14.577 & 9.693 & \textbf{9.230} \\

\midrule

\textit{PG-19} \\ 
Full Cache & 23.787 & 23.787 & 23.787 \\
\texttt{Baseline} & 37.013 & 27.939 & 25.451 \\
\texttt{Baseline+} & 35.832 & 27.829 & 25.429\\
$\hho$+Performer & 36.996 & 27.938 & 25.451 \\
\methodname{} (2\%) & \textbf{32.157} & 27.887 & 26.322 \\
\methodname{} (5\%) & 33.195 & \textbf{27.089} & 25.979 \\
\methodname{} (10\%)  & 41.204 & 27.201 & \textbf{25.134}\\
\bottomrule
\end{tabular}
\end{sc}
\end{small}
\end{center}
\label{tab:llama_lm}
\end{table}

Evaluation results \cite{ clark2019boolq, cui2020mutual} with $\Lambda$-masking in Table \ref{tab:llama_lm_lambda} show \methodname{}'s benefit to a different sparse policy (though less performative than $\hho$). Similar to the case with $\hho$, \methodname{} closes the gap from full caching but cannot match the performance completely. TOVA also observes similar benefits in language modeling, shown Appendix~\ref{app:lm}. While \methodname{} is efficacious for language modeling and classification, we also want to assess its utility for generation where the KV cache storage becomes a critical bottleneck.

\begin{table}[t]
\caption{Llama 2 7B performance on WikiText (word perplexity), MuTual (16-shot R@1), and  BoolQ (10-shot accuracy) with 5\% $\Lambda$-masking as the primary underlying sparse policy.}
\begin{center}
\begin{small}
\begin{sc}
\begin{tabular}{lcccc}
\toprule
$\Lambda$ Method & WikiText ($\downarrow$) & MuTual & BoolQ \\
\midrule
Full Cache & 8.79 & 55.08 & 80.40 \\
\texttt{Baseline}   & 10.66 & 53.50 & 77.28\\
\texttt{Baseline+}  & 10.64 & 53.27 & 77.46 \\
\methodname{} (5\%) & \textbf{10.12} & \textbf{54.51} & \textbf{78.65}\\
\bottomrule
\end{tabular}
\end{sc}
\end{small}
\end{center}
\label{tab:llama_lm_lambda}
\end{table}

\subsection{Summarization}
\label{sec:summarization}

Now, we move on to generation, specifically summarization, to test the ability to generate longer and coherent sequences by synthesizing numerous tokens. Unlike in our language modeling evaluations, the model will have access to all tokens during the prompting phase with the sparse policy and \methodname{} only kicking in during the subsequent generation steps. Consequently, generation sparse policies are fundamentally different from the language modeling masks \methodname{} is trained on, yet despite this, we show that our method maintains its superior performance.

\begin{table}[t]
\caption{Llama 2 13B and Falcon 7B generation quality comparison on CNN/DailyMail and XSum with 408 sparse tokens (10\% and 20\% of the context lengths of Llama 2 and Falcon, respectively) with $\hho$ as the primary underlying sparse policy. Llama 2 13B is given 5 shots while Falcon 7B is given 3 shots due to its shorter context length. Values are in the format [Rouge-1/2/L].}
\begin{center}
\begin{small}
\begin{sc}
\begin{tabular}{lcc}
\toprule
$\hho$ Method & CNN/DailyMail & XSum\\
\midrule

\textit{Llama 2 13B} \\
Full Cache & 27.55/9.96/25.80 & 33.14/13.05/27.33 \\
\texttt{Baseline}  & 23.57/7.35/22.04 & 33.09/\textbf{13.09}/\textbf{27.44} \\
\texttt{Baseline+}  & 23.40/7.31/21.88 & 33.09/13.06/27.41\\
\methodname{} (2\%) & \textbf{25.27}/\textbf{7.76}/\textbf{23.64} & \textbf{33.40}/12.98/27.41 \\
\methodname{} (5\%) & 24.45/7.70/22.87 & 33.15/13.02/27.39\\

\midrule

\textit{Falcon 7B} \\
Full Cache & 25.92/8.52/24.15 & 27.17/8.83/22.67 \\
\texttt{Baseline}  & 21.26/5.95/19.73 & 24.50/7.65/20.50 \\
\texttt{Baseline+}  & 21.31/6.16/19.75 & 24.55/7.66/20.56\\
\methodname{} (5\%) & 23.00/6.28/21.28 & 24.94/8.17/20.94\\
\methodname{} (10\%) & \textbf{23.22}/\textbf{6.37}/\textbf{21.53} & \textbf{25.21}/\textbf{8.28}/\textbf{21.17} \\
\bottomrule
\end{tabular}
\end{sc}
\end{small}
\end{center}
\label{tab:h2o_gen}
\end{table}

\begin{table}[t]
\caption{Llama 2 7B performance on MultiNews (1-shot), CNN/DailyNews (5 shot), and XSum (5-shot) with 5\% and 10\% $\hho$ as the primary underlying test sparse policies. Values are in the format [Rouge-1]/[Rouge-2]/[Rouge-L].}
\begin{center}
\begin{small}
\begin{sc}
\begin{tabular}{lcc}
\toprule
$\hho$ Method & 5\% $\hho$ & 10\% $\hho$ \\
\midrule
\textit{MultiNews} \\
Full Cache & 23.79/6.87/21.35 & 23.79/6.87/21.35 \\
\texttt{Baseline}  & 13.38/3.25/12.25 & 19.44/4.97/17.73 \\
\texttt{Baseline+}  & 13.58/3.32/12.41 & 19.44/4.96/17.72 \\
\methodname{} (2\%) & 15.31/3.73/14.03 & 20.32/5.24/18.51  \\
\methodname{} (5\%) & \textbf{15.42}/\textbf{3.80}/\textbf{14.14} & \textbf{20.55}/\textbf{5.29}/\textbf{18.70} \\

\midrule
\textit{CNN/DailyMail} \\
Full Cache & 26.25/9.34/24.40 & 26.25/9.34/24.40 \\
\texttt{Baseline}  & 18.18/4.92/16.89 & 20.04/6.09/18.66 \\
\texttt{Baseline+}  & 18.24/4.91/16.85 & 20.15/6.21/18.73 \\
\methodname{} (2\%) & 18.71/5.40/17.34 & 20.76/6.40/19.32 \\
\methodname{} (5\%) & \textbf{19.21}/\textbf{5.44}/\textbf{17.80} & \textbf{22.29}/\textbf{6.85}/\textbf{20.69} \\

\midrule
\textit{XSum} \\
Full Cache & 30.65/11.11/25.40 & 30.65/11.11/25.40 \\
\texttt{Baseline} & 29.03/10.77/24.28 & 30.68/\textbf{11.54}/25.58 \\
\texttt{Baseline+}  & 28.94/10.78/24.15 & 30.64/11.49/\textbf{25.59} \\
\methodname{} (2\%) & \textbf{30.72}/\textbf{11.53}/\textbf{25.57} & 30.34/10.98/25.31 \\
\methodname{} (5\%) & 30.03/11.19/25.03 & \textbf{30.82}/11.17/25.56 \\

\bottomrule
\end{tabular}
\end{sc}
\end{small}
\end{center}
\label{tab:h2o_llama_gen}
\end{table}

In Tables \ref{tab:h2o_gen} and \ref{tab:h2o_llama_gen}, we see \methodname{} achieves better ROUGE \cite{lin-2004-rouge} scores than purely $\hho$ on the CNN/DailyMail \cite{hermann2015teaching, see-etal-2017-get}, MultiNews \cite{alex2019multinews}, and XSum \cite{narayan2018don} datasets. Even at exceptionally low sparsity levels, $\hho$ can capture a significant amount of the full cache's performance. This is even more surprising for Falcon models which already cache many times fewer tokens than Llama 2 due to the multi-query attention mechanism. Despite this, we observe \methodname{} surpasses the already strong performance of $\hho$ across the board where $\hho$ underperforms compared to the full cache. Like in language modeling, we again see that the improvement from \texttt{Baseline} to \texttt{Baseline+} pales in comparison to the improvement induced by \methodname{}, sometimes even matching the full cache performance as in XSum. Again, we also see the transferability of \methodname{} to other sparsity levels. See Appendix~\ref{app:gen_outputs} for example generation outputs.



\subsection{Latency and Throughput}
\label{sec:efficiency}

\begin{table}[t]
\caption{Llama 2 7B and 13B's generation throughput (tokens/s) and latency (s) on an A100 GPU. In the sequence length column, we use "5000 + 5000"
to denote a prompt length of 5000 and a generation length of 5000. "OOM" stands for out-of-memory.}
\begin{center}
\begin{small}
\begin{sc}
\begin{tabular}{ccccccc}
\toprule
Seq. length & Model size & Batch size & Metric & Full Cache & \texttt{Baseline+} & \methodname{} (5\%) \\\midrule
5000+5000 &  13B & 4 & latency & 257.3 & 185.2 & 204.7  \\
2048+2048 &  7B & 24 & latency & 116.7 & 78.3 & 95.1\\
\midrule
2048+2048 &  7B & 24 & throughput  & 421.2 & 627.7 & 516.9\\
2048+2048 &  7B & 64 & throughput  & OOM & 819.2 & 699.2\\
\bottomrule
\end{tabular}
\end{sc}
\end{small}
\end{center}
\label{efficient_metric}
\end{table}

Following Sheng et al. \cite{Sheng2023HighthroughputGI}, we benchmark the generation throughput and latency of \methodname{} on an NVIDIA A100 80G GPU using FP16 precision. We focus on the Llama 2 7B and 13B models, with all speedup results tested end-to-end with both prompting and generation phases. To measure its performance when generating long sequences or inputting large batch sizes, we use synthetic datasets where all prompts are padded to the same length and batched together. The same number of tokens are generated for each prompt. We test different combinations of prompt and generation lengths. 

Table~\ref{efficient_metric} shows results with sequence lengths from 4K to 10K. With the same batch size, \methodname{} reduces the latency by $1.1 - 1.3\times$ compared to the full cache, though slightly slower than $\hho$. Moreover, \methodname{} saves memory to allow larger batch sizes with a $1.7\times$ improvement on generation throughput for Llama 2 7B, closely matching the performance of $\hho$. 


\subsection{Empirical Analysis and Ablations}
\label{sec:ablation}

Now that we have shown that \methodname{} is simple and effective, we share some interesting characteristics of our method.

\vspace{-0.1in}
\paragraph{Reconstructing Attention Probabilities.}

Sparse KV cache policies can delete tokens that may be needed later on. A way to see this is to construct the sparse attention matrix and compare with  the full one. In Figure~\ref{fig:attn_maps},  $\hho$ zeroes out many relatively high attention probabilities with a bias towards keeping early tokens. More examples are in Appendix~\ref{app:attn_visualization}. Visually, \methodname{} provides a sketch of the deleted tokens which appears to reasonably reconstruct trends.

\begin{figure}[h]
\centering 
\includegraphics[width=0.45\columnwidth]{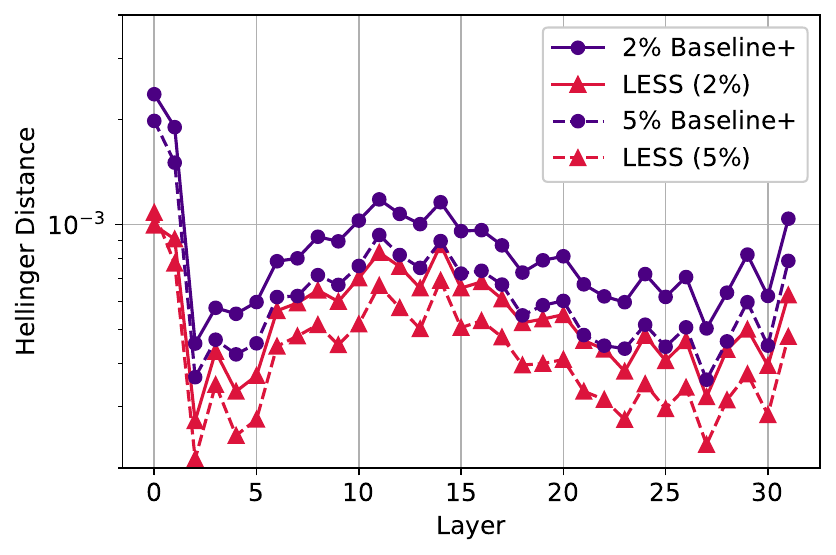}
\caption{Layer-wise Llama 2 7B mean Hellinger distance from original attention probabilities, aggregated across WikiText evaluation samples. The underlying sparse policy is $\hho$. Here, \methodname{} is evaluated based on their training sparsity percentages.}
\label{fig:layer_vs_prob_dist}
\end{figure}

Numerically, we measure the similarity of each row in the attention matrix with corresponding rows produced by $\hho$ and \methodname{} with the Hellinger distance, which for two discrete probability vectors, $\bp$ and $\bq$, is defined as
\begin{align}
    \cH(\bp, \bq) \coloneqq \| \sqrt{\bp} - \sqrt{\bq} \|_2 / \sqrt{2}
\end{align}
where the square root is elementwise. The value of $\cH(\bp, \bq)$ ranges from 0 to 1, where a lower value indicates greater similarity. In Figure~\ref{fig:layer_vs_prob_dist}, we see that our method more accurately replicates the original attention probability distributions as measured by the Hellinger distance. We choose to aggregate each layer separately since the attention distribution patterns tend to vary dramatically throughout the model. 


\paragraph{Larger Kernels.}

\begin{figure}[ht]
\begin{center}
\centerline{\includegraphics[width=0.45\columnwidth]{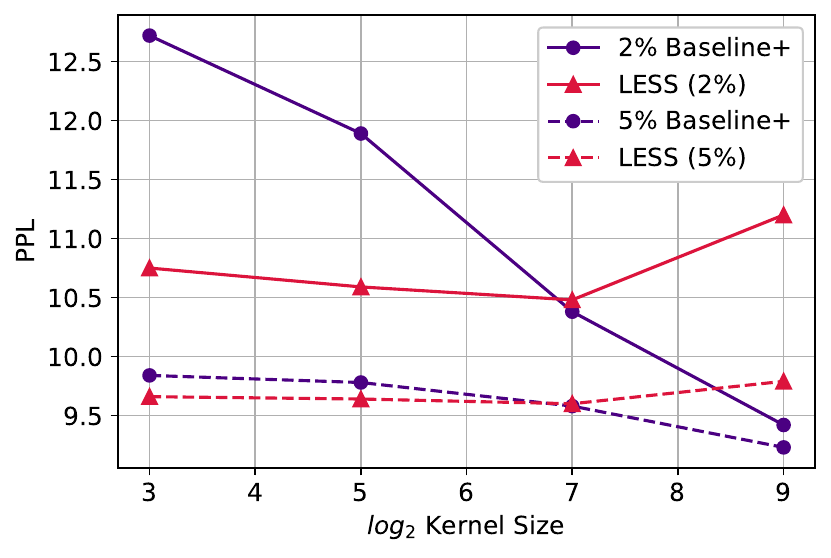}}
\caption{Llama 2 7B WikiText word perplexity (lower is better) as the kernel size quadruples, compared against \texttt{Baseline+} which occupies the same space. The sparse KV cache policy is $\hho$.
}
\label{kernel_vs_ppl}
\end{center}
 \vskip -0.2in
\end{figure}

In our experiments, we fixed $R = 8$, and as we show in Figure \ref{kernel_vs_ppl}, the performance generally increases as $R$ increases. However, at a certain point, the marginal benefit derived from increasing $R$ is less than shifting more of the KV cache to the sparse policy, suggesting that a small low-rank cache is enough.

\paragraph{Providing Hope for Long Sequences.}

\begin{figure}[h]

\centering
\includegraphics[width=0.4\columnwidth]{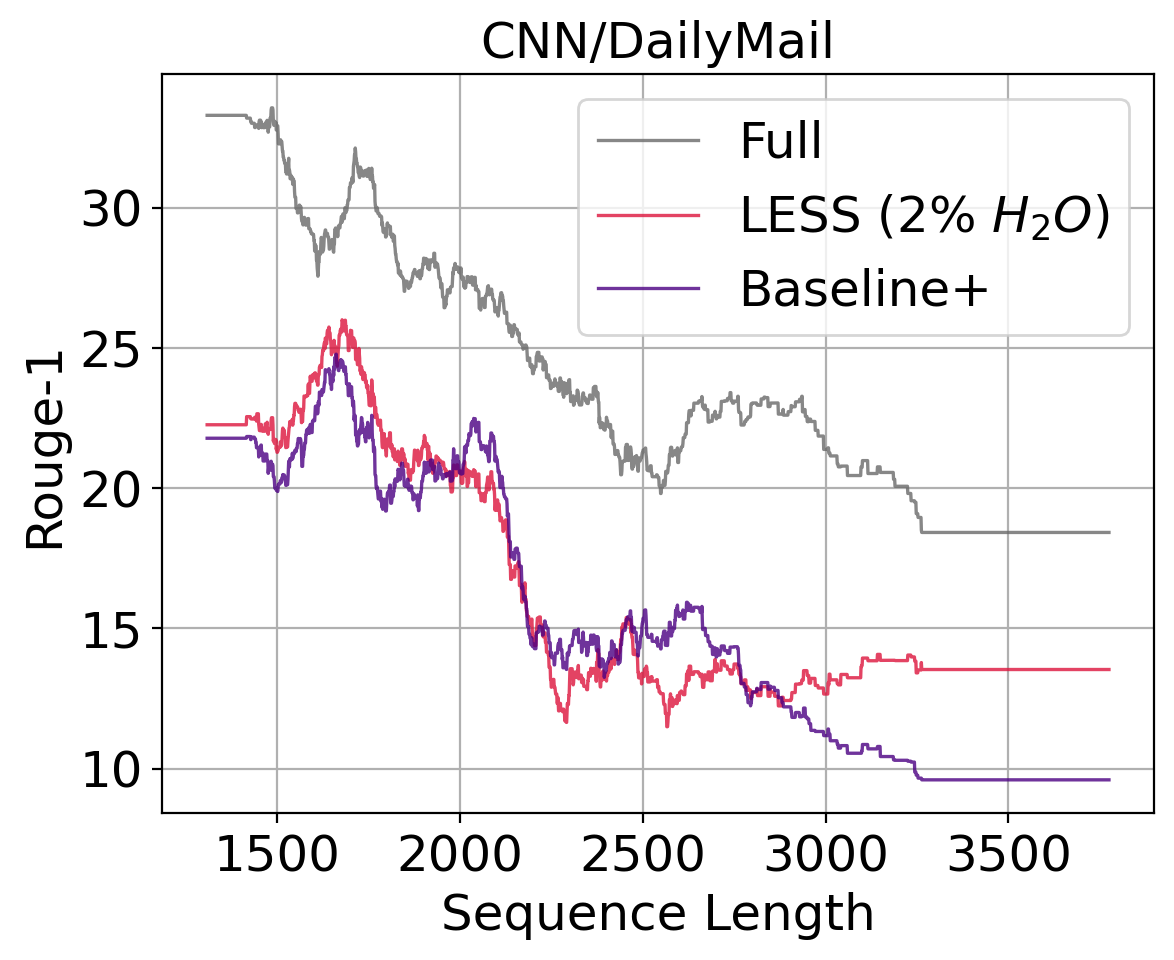}
\includegraphics[width=0.43\columnwidth]{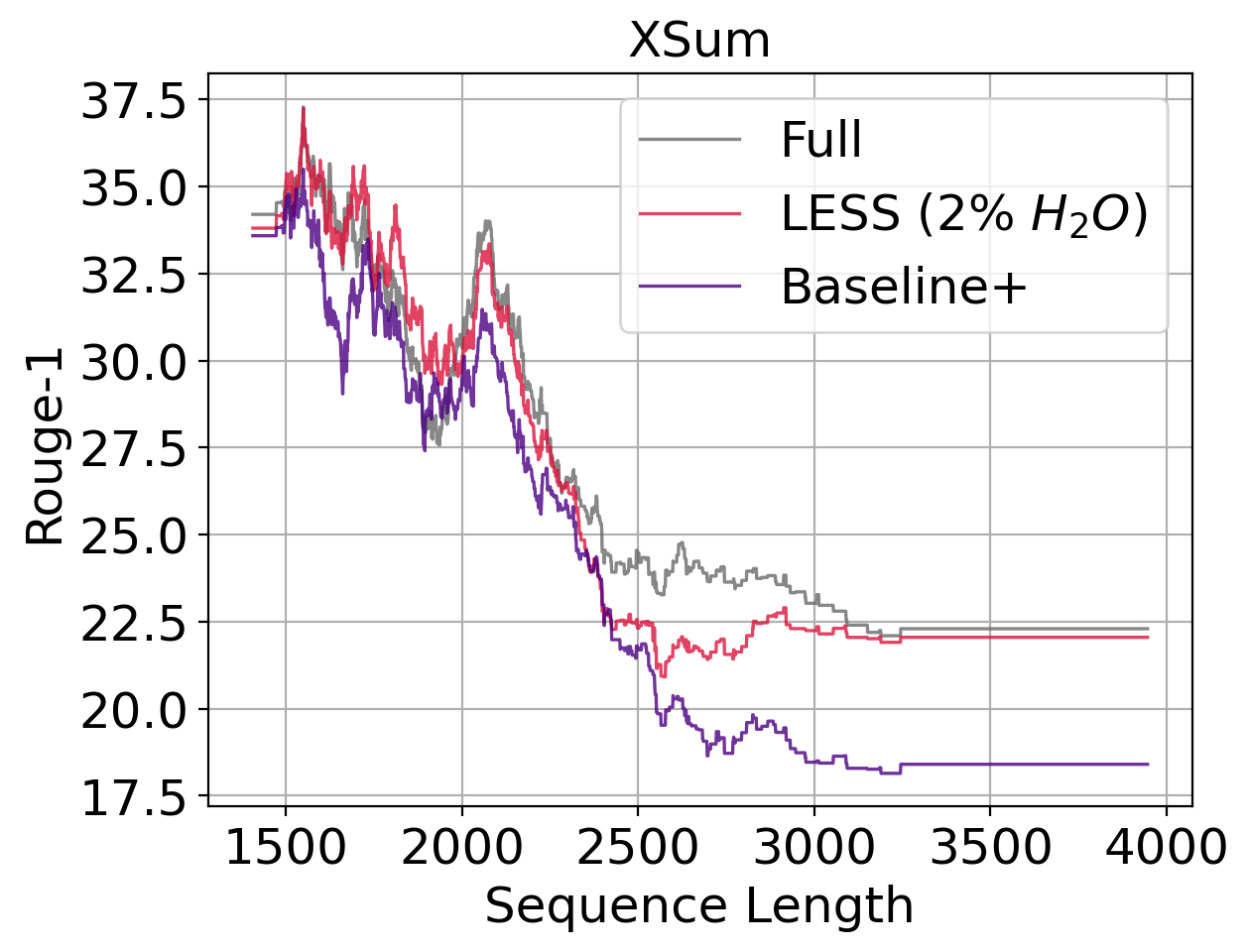}
\vskip -0.1in
\caption{Relationship between Rouge-1 score and prompt length for Llama 2 7B with different cache methods on CNN/DailyMail (left) and XSum (right). The test sparse KV cache policy is 5\% $\hho$ for all models. As these results can be fairly noisy, the lines are $k$-nearest regression lines where $k$ is 10\% of the dataset size.}
\label{fig:len_vs_gen}
\vskip -0.1in
\end{figure}

Model performance appears to be highly correlated with the input sequence length regardless of the caching method. As shown in Figure~\ref{fig:len_vs_gen}, even the full cache model performance drops dramatically and immediately as the prompt length increases. \texttt{Baseline+} and  \methodname{} (1\% $\hho$) appear to perform similarly for shorter sequences but diverge for longer sequences where we see \methodname{} is more performative. This follows our intuition since for sparse cache policies, a smaller fraction of KV pairs is saved as the sequence length increases, so more information is omitted. This is where a low-rank state can help to recover some of this information.

\section{Conclusion and Future Work}
\label{sec:conclusion}

To tackle the KV cache bottleneck, we introduce \methodname{} which has demonstrated itself to be an effective way to boost eviction-based KV cache algorithms. Motivated by the necessity to maintain information that would have been discarded, the constant-sized \methodname{} recovers a significant portion of the performance lost due to maintaining a small cache across a variety of scenarios and intensities, despite being cheap to train and deploy. There are many exciting avenues of work that can enhance \methodname{} or build upon it, such as improving kernel design and investigating the residual of \methodname{}. Such directions will further push the performance of a condensed KV cache to that of a complete cache, allowing LLMs to accomplish the same tasks with less.

\section*{Acknowledgements}

The work of H. Dong is supported in part by the Liang Ji-Dian Graduate Fellowship, the Michel and Kathy Doreau Graduate Fellowship in Electrical and Computer Engineering, and the Wei Shen and Xuehong Zhang Presidential Fellowship at Carnegie Mellon University. Z. Wang is in part supported by a Google Research Scholar Award and the NSF AI Institute for Foundations of Machine Learning (IFML). The work of Y. Chi is supported in part by the grants NSF DMS-2134080 and ONR N00014-19-1-2404. B. Chen is supported in part by MOFFETT AI gift funding.

\bibliographystyle{alphaabbr}
\bibliography{bibliography.bib}

\newpage

\appendix

\section{Attention Matrix Visualizations}
\label{app:attn_visualization}

\begin{figure}[h]
    \begin{center}
    \centering
    \includegraphics[width=0.7\columnwidth]{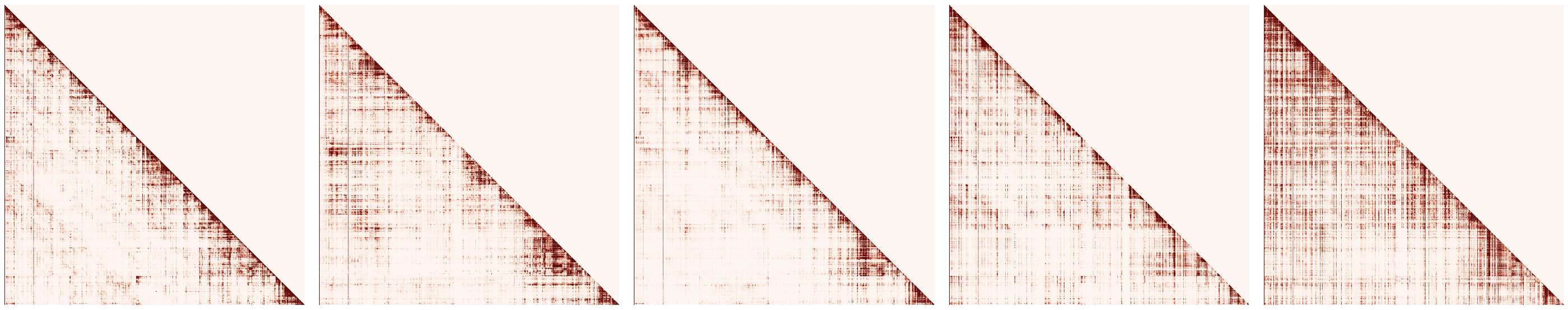} \\
    \includegraphics[width=0.7\columnwidth]{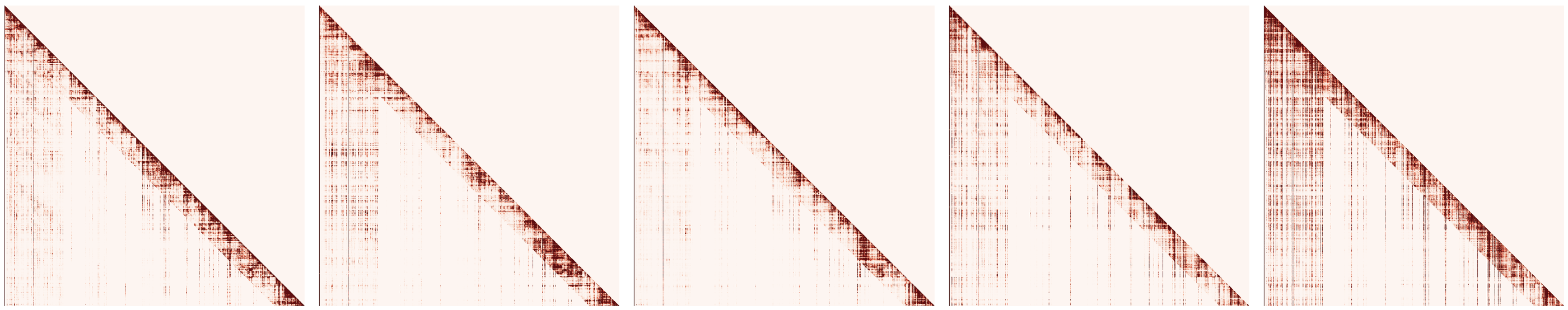} \\
    \includegraphics[width=0.7\columnwidth]{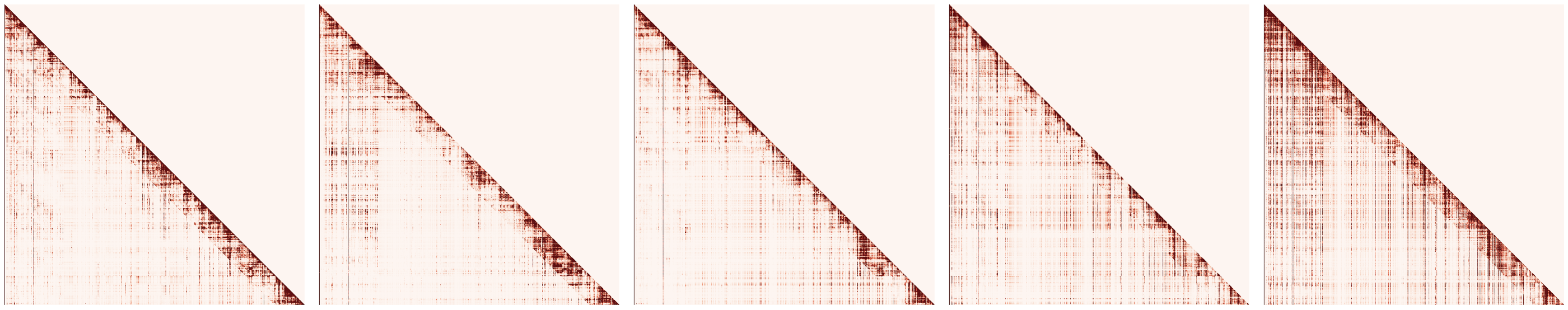} \\
    \caption{Example attention probability matrices from passing a single input into Falcon 7B. From top to bottom, the rows consist of attention maps from the original model, 10\% $\hho$ (204 tokens), and \methodname{} (10\% $\hho$). Darker pixels indicate larger probability weights. Only the first 1024 tokens are displayed.}
    \label{fig:falcon_h2o_attn_maps}
    \end{center}
    \vskip -0.2in
\end{figure}

\begin{figure}[ht]
    \begin{center}
    \centering
    \includegraphics[width=0.7\columnwidth]{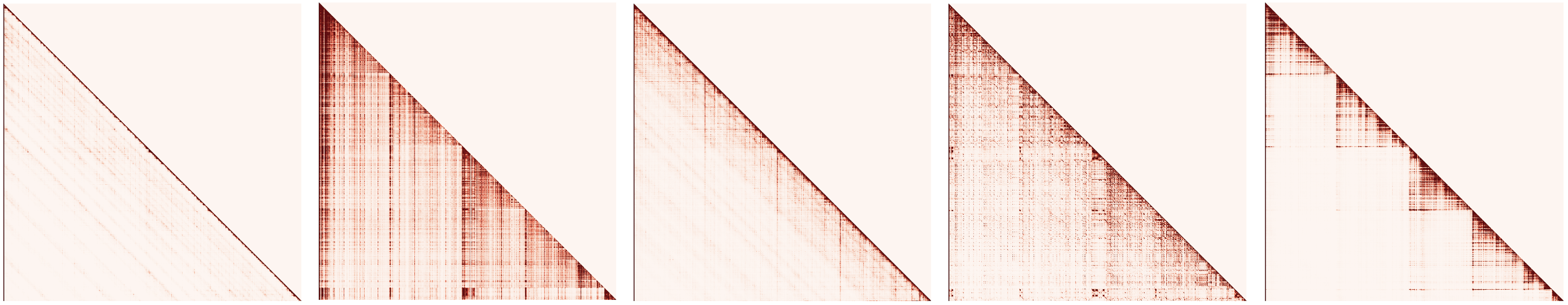} \\
    \includegraphics[width=0.7\columnwidth]{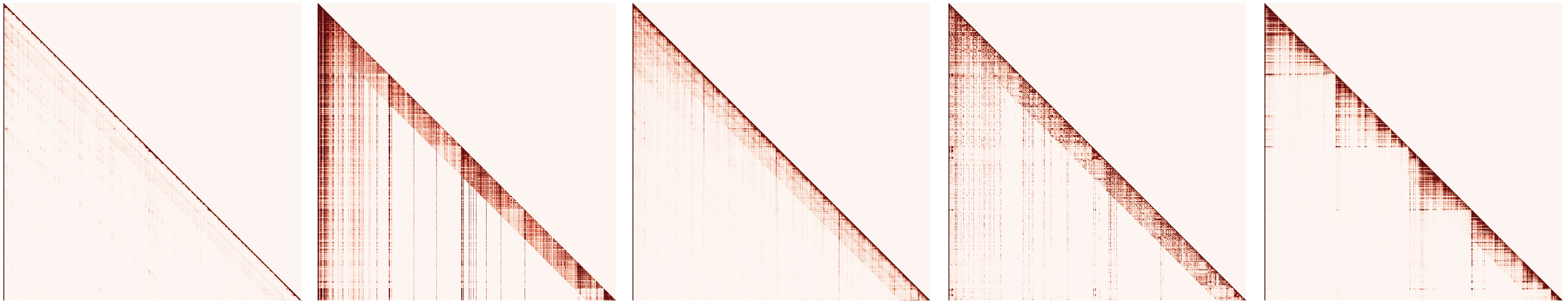} \\
    \includegraphics[width=0.7\columnwidth]{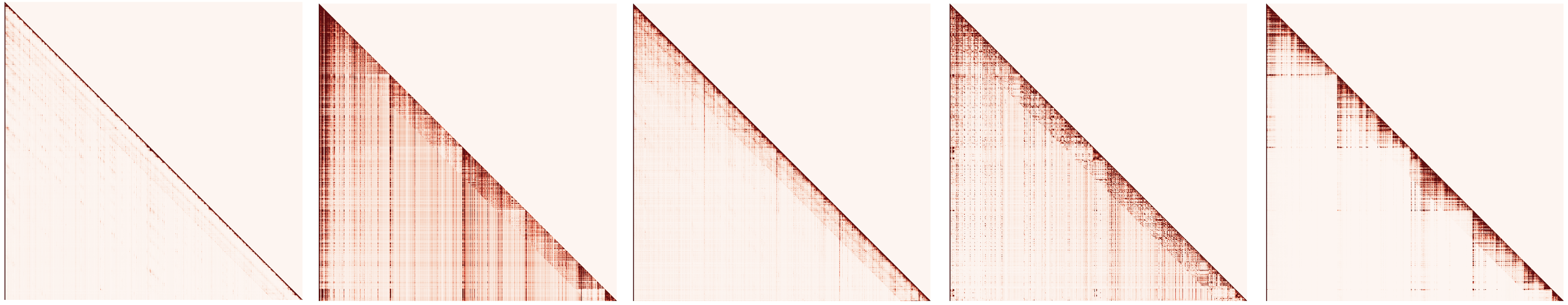} \\
    \caption{Example attention probability matrices from passing a single input into Llama 2 7B. From top to bottom, the rows consist of attention maps from the original model, 5\% $\hho$ (204 tokens), and \methodname{} (5\% $\hho$). Darker pixels indicate larger probability weights. Only the first 1024 tokens are displayed.}
    \label{fig:llama_h2o_attn_maps}
    \end{center}
    \vskip -0.2in
\end{figure}

\begin{figure}[ht]
    \begin{center}
    \centering
    \includegraphics[width=0.7\columnwidth]{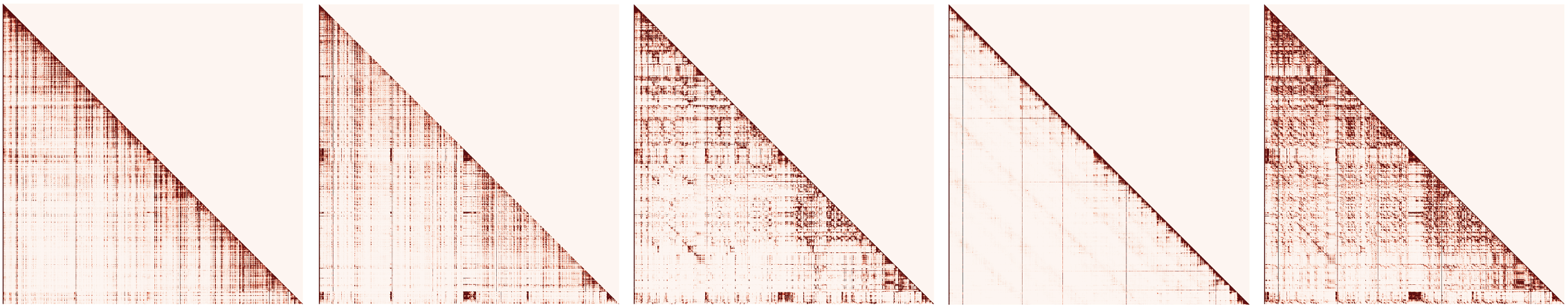} \\
    \includegraphics[width=0.7\columnwidth]{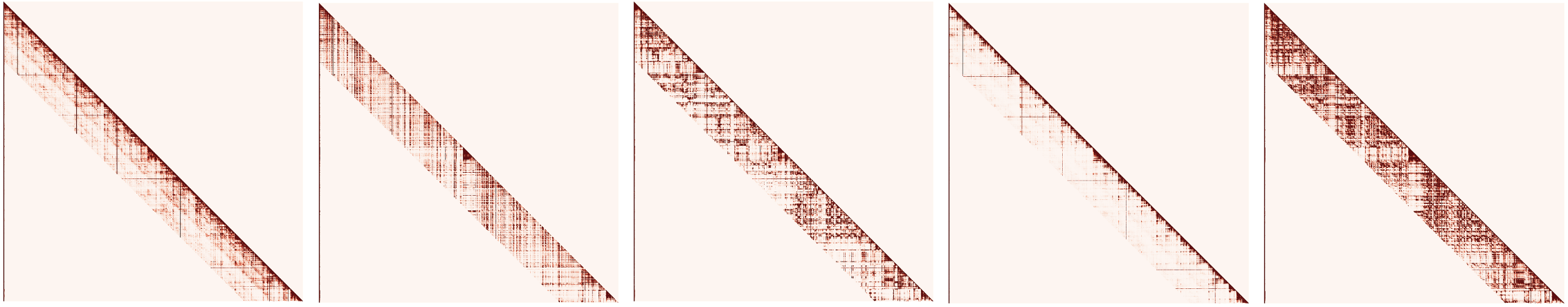} \\
    \includegraphics[width=0.7\columnwidth]{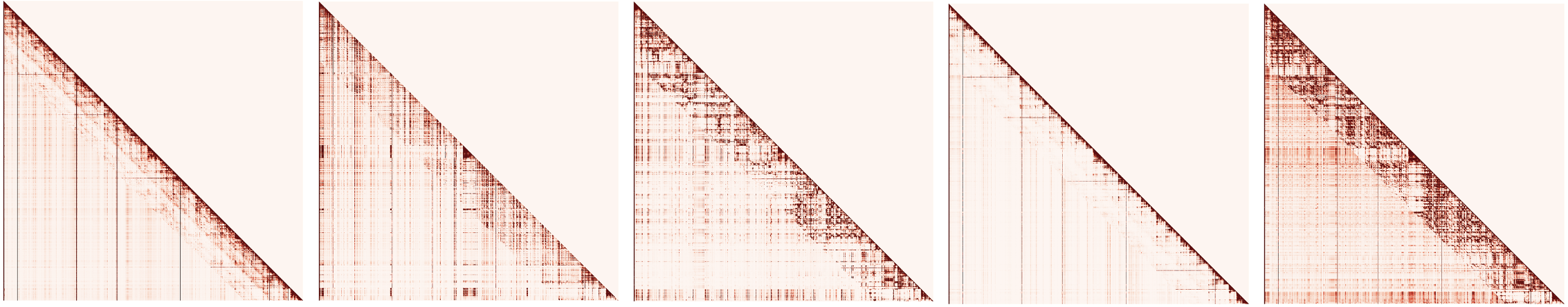} \\
    \caption{Example attention probability matrices from passing a single input into Llama 2 7B. From top to bottom, the rows consist of attention maps from the original model, 5\% $\Lambda$-masking (204 tokens), and \methodname{} (5\% $\Lambda$). Darker pixels indicate larger probability weights. Only the first 1024 tokens are displayed.}
    \label{fig:llama_stream_attn_maps}
    \end{center}
    \vskip -0.2in
\end{figure}

This section provides some qualitative results on attention matrix approximations by sparse policies and \methodname{}. While low-rank caches \methodname{} cannot perfectly recover all the missing information, it visually is able to reconstruct a patterns that are completely ignored by sparse policies. We can also see the idiosyncrasies of the sparse policies and \methodname{}, such as $\hho$'s bias towards keeping early tokens, as shown in Figures~\ref{fig:falcon_h2o_attn_maps} and \ref{fig:llama_h2o_attn_maps}, and $\Lambda$-masking's tendency to miss influential tokens which are captured by \methodname{}, as show in Figure~\ref{fig:llama_stream_attn_maps}.

\section{Language Modeling Comparison}
\label{app:lm}

In Table~\ref{tab:lm_comparison}, we compare language modeling using \methodname{} with $\hho$, $\Lambda$-masking, and TOVA. We see that \methodname{} makes a greater improvement on word perplexity than simply caching the equivalent number of extra tokens, though none of the methods can fully recover the original model's performance.

\begin{table}[h]
\caption{Llama 2 7B performance (word PPL) on WikiText and PG-19 at 5\% token sparsity using different sparse policies. Lower is better.}
\begin{center}
\begin{small}
\begin{sc}
\begin{tabular}{lccccccc}
\toprule 
Method & $\hho$ & $\Lambda$ & TOVA \\
\midrule
\textit{WikiText} \\
Full Cache & 8.79 & 8.79 & 8.79 \\
\texttt{Baseline}   & 9.86 & 10.66 & 9.97 \\
\texttt{Baseline+}  & 9.84 & 10.64 & 9.95 \\
\methodname{} (5\%) & \textbf{9.66} & \textbf{10.12} & \textbf{9.72} \\

\midrule
\textit{PG-19} \\
Full Cache & 23.79 & 23.79 & 23.79 \\
\texttt{Baseline}   & 27.94 & 22K & 27.88 \\
\texttt{Baseline+}  & 27.83 & 22K & 27.79 \\
\methodname{} (5\%) & \textbf{27.09} & \textbf{3.9K} & \textbf{27.34} \\
\bottomrule
\end{tabular}
\end{sc}
\end{small}
\end{center}
\label{tab:lm_comparison}
\end{table}

\section{Latency Breakdown}
\label{app:breakdown}

Here, we show that maintaining and using the low-rank state in \methodname{} has little overhead by determining the latencies of different operations in \methodname{}. Shown in Table~\ref{tab:breakdown}, the total latency of \methodname{} is much smaller than the full cache but slightly higher than \texttt{Baseline+} due to the additional operations associated with the low-rank state. This overhead represents about 15\% of the decoding time, meaning it does not significantly impact the overall efficiency.

\begin{table*}[h]
\caption{Latency (s) breakdown for Llama 2 7B on an A100 GPU, with a batch size of 64, prompt length of 512, and generation length of 512 with 5\% $\hho$ as the underlying sparse policy. ``Eviction'' refers to $\hho$'s KV eviction algorithm, ``Kernels'' refers to \eqref{eq:kernel_q} and \eqref{eq:kernel_k}, ``Attention Synthesis'' refers to \eqref{eq:attn_approx}, and ``LR Cache Update'' refers to \eqref{eq:h_update} and \eqref{eq:z_update}.}
\begin{center}
\begin{small}
\begin{sc}
\begin{tabular}{lcccccc}
\toprule
        & Decoding & Eviction & Kernels & Attention Synthesis & LR Cache Update & Total \\\midrule
Full Cache  & 50.71 & N/A & N/A & N/A & N/A  & 50.71\\ 
\texttt{Baseline+}  & 23.53 & 4.52 & N/A & N/A & N/A  & 28.05\\ 
\methodname{} (5\%)  & 23.61 & 4.39 & 0.87 & 1.35 & 1.51  & 32.96\\ 
\bottomrule
\end{tabular}
\end{sc}
\end{small}
\end{center}
\label{tab:breakdown}
\end{table*}

\newpage

\section{Generation Outputs}
\label{app:gen_outputs}

\begin{figure}[h]
    \begin{center}
    \centering
    \includegraphics[width=0.95\columnwidth]{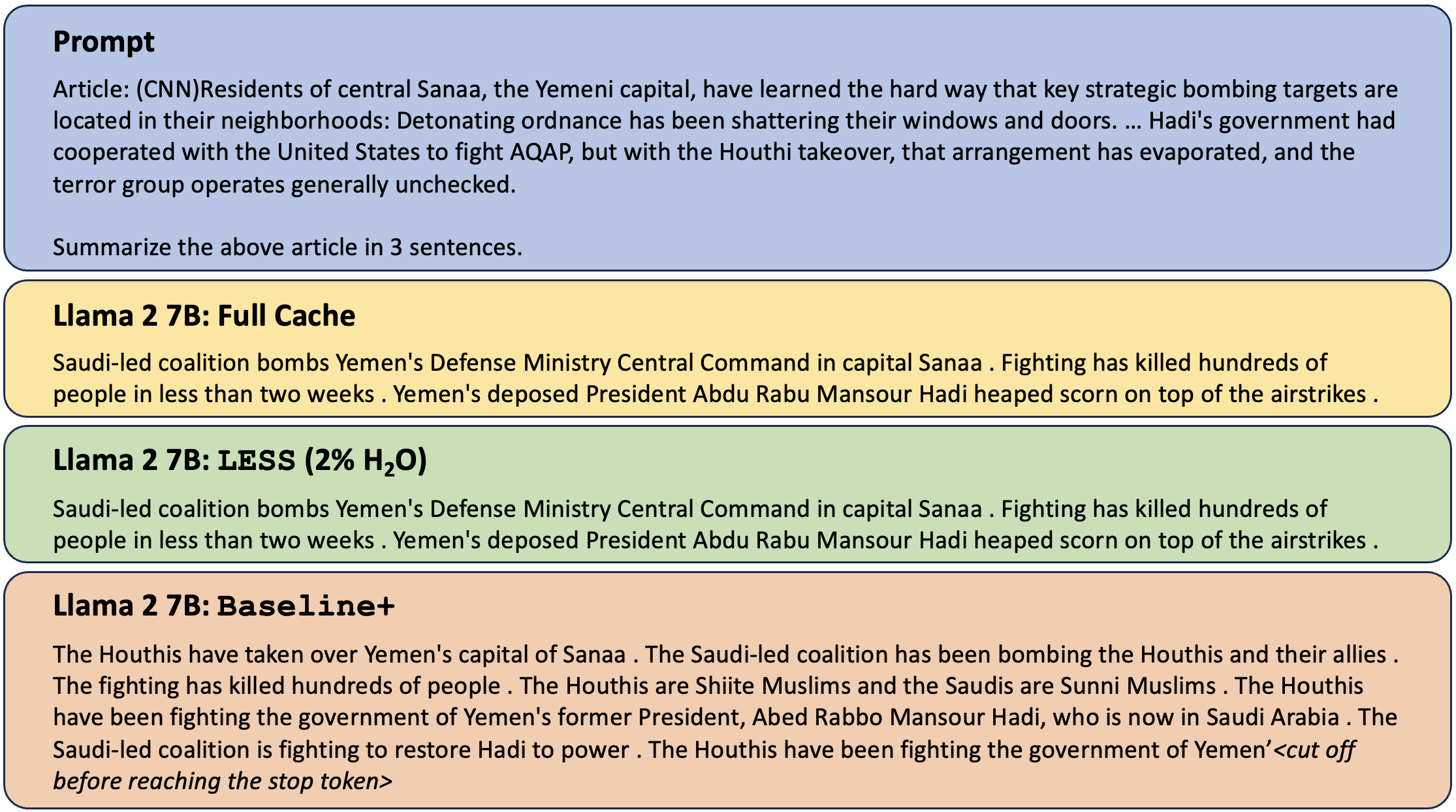} \\
    
    \caption{Example 5-shot (not shown) CNN/DailyMail summary generation results produced by variations of Llama 2 7B with an underlying sparse policy of 2\% $\hho$. For brevity, only the start and end of the article are shown with the middle omitted with an ellipsis. \methodname{} produces the same concise summary as the full cache while \texttt{Baseline+} produces rambling text, exceeding the 3 sentence requirement by the prompt. The original article is from \cite{brumfield2015death}.}
    \label{fig:example_llama2_out1}
    \end{center}
    \vskip -0.2in
\end{figure}

\begin{figure}[h]
    \begin{center}
    \centering
    \includegraphics[width=0.95\columnwidth]{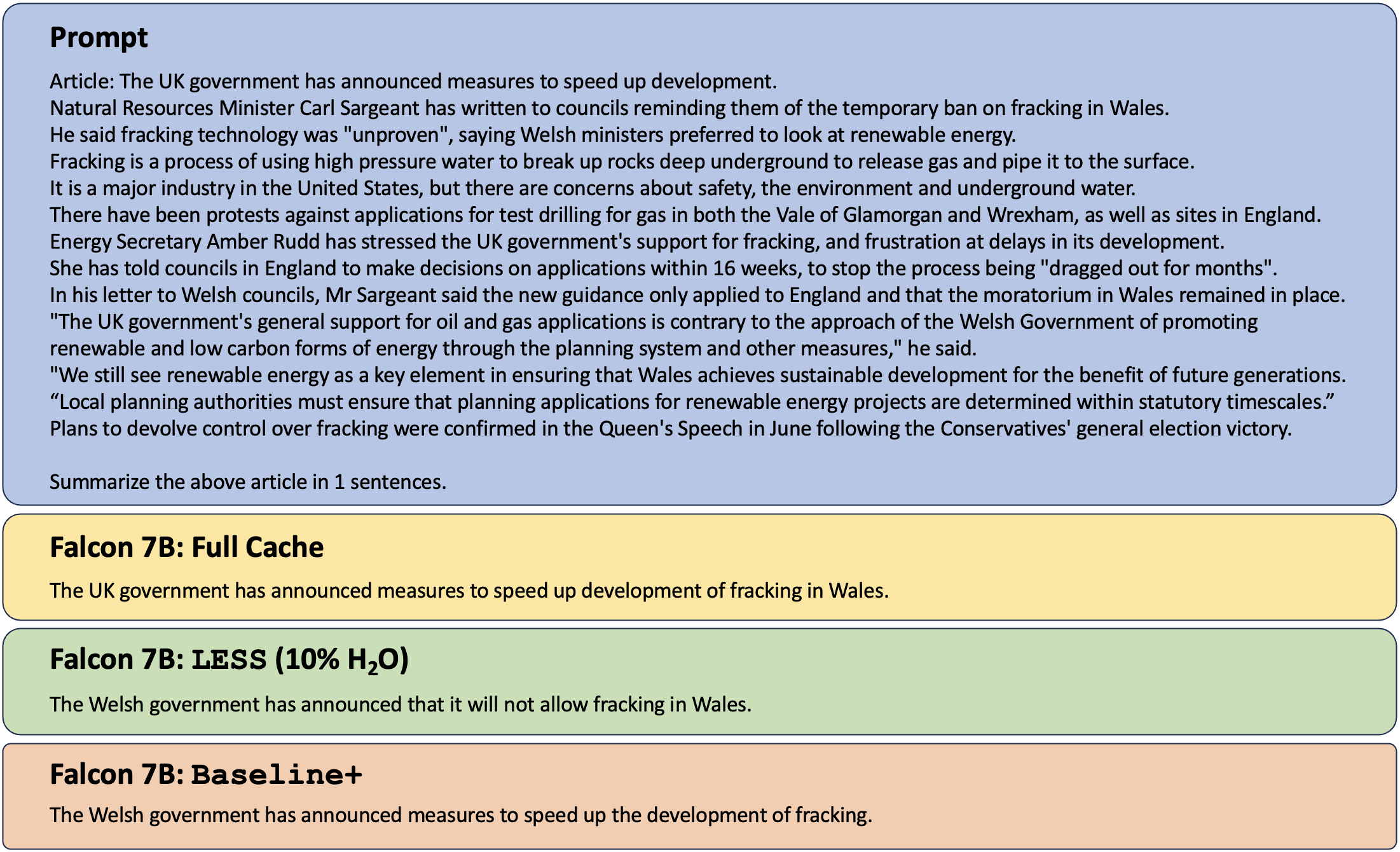} \\
    \caption{Example 3-shot (not shown) XSum summary generation results produced by variations of Falcon 7B. Models were evaluated with 20\% $\hho$. The summary by \texttt{Baseline+} is factually incorrect based on the article, while \methodname{} preserves the meaning better. The original article is from \cite{bbc2015fracking}.}
    \label{fig:example_falcon_out1}
    \end{center}
    \vskip -0.2in
\end{figure}

We include a couple examples of generation outputs in Figure~\ref{fig:example_llama2_out1} and Figure~\ref{fig:example_falcon_out1}. In both cases, the full cache, \methodname{}, and \texttt{Baseline+} models attempt to summarize news articles. We see in Figure~\ref{fig:example_llama2_out1} that \methodname{} is able to produce the same concise summary as the full cache while \texttt{Baseline+} produces rambling text. In Figure~\ref{fig:example_falcon_out1}, we observe that \methodname{} completely changes the meaning of the summary from $\hho$ alone--\texttt{Baseline+} is factually incorrect based on the article.

\end{document}